%% file: main.tex
\input{preamble.tex}

\mlsystitlerunning{Cross-Record Precision Routing and Sensitivity Inflation}
\hypersetup{pdftitle={Sensitivity Inflation from Cross-Record Precision Routing in Private Quantized Fine-Tuning},pdfsubject={MLSys submission},pdfauthor={}}
\begin{document}
\twocolumn[
\mlsystitle{Sensitivity Inflation from Cross-Record Precision Routing in Private Quantized Fine-Tuning}
\begin{mlsysauthorlist}
\mlsysauthor{Annan Zhang}{pku}
\mlsysauthor{Shengjie Song}{pku}
\mlsysauthor{Wenshuai Yao}{pku}
\mlsysauthor{Tianmiao}{buaa}
\mlsysauthor{Lihao Zhang}{pku}
\end{mlsysauthorlist}
\mlsysaffiliation{pku}{Peking University}
\mlsysaffiliation{buaa}{Beihang University}
\mlsyscorrespondingauthor{Annan Zhang}{\mbox{annan@pku.edu.cn}}
\vskip 0.3in
\begin{abstract}
\input{abstract}
\end{abstract}
]
\printAffiliationsAndNotice{}
\input{body}
\bibliography{references}
\bibliographystyle{mlsys2025}
\clearpage
\appendix
\input{appendix}
\end{document}

%% file: preamble.tex
\documentclass{article}
\usepackage{mlsys2025}
\renewcommand{\icmlruler}[1]{}
\usepackage{etoolbox}
\makeatletter
\patchcmd{\mlsysauthor}{\ifdefined\isaccepted}{\iftrue}{}{%
  \PackageError{authordetails}{Could not show author names}{Check the MLSys style.}}
\patchcmd{\mlsysaffiliation}{\ifdefined\isaccepted}{\iftrue}{}{%
  \PackageError{authordetails}{Could not show affiliations}{Check the MLSys style.}}
\patchcmd{\mlsyscorrespondingauthor}{\ifdefined\isaccepted}{\iftrue}{}{%
  \PackageError{authordetails}{Could not show the corresponding author}{Check the MLSys style.}}
\makeatother
\usepackage[T1]{fontenc}
\usepackage[utf8]{inputenc}
\usepackage{microtype}
\usepackage{hyperref,url}
\usepackage{xurl}
\usepackage{amsmath,amssymb,mathtools,amsthm}
\usepackage{booktabs,tabularx,array,multirow}
\usepackage{graphicx,xcolor}
\usepackage{enumitem}
\usepackage{algorithm,algorithmic}
\renewcommand{\algorithmicrequire}{\textbf{Input:}}
\renewcommand{\algorithmiccomment}[1]{\hfill{\color{ink2}\(\triangleright\)~#1}}
\newcolumntype{Y}{>{\raggedright\arraybackslash}X}

\usepackage{placeins}

\theoremstyle{plain}
\newtheorem{proposition}{Proposition}
\input{figures/figstyle}

%% file: figures/figstyle.tex
\usepackage{pgfplots}
\pgfplotsset{compat=1.18}
\usetikzlibrary{arrows.meta,calc,positioning,backgrounds,decorations.pathreplacing}
\usepgfplotslibrary{groupplots,fillbetween}

\definecolor{native}{HTML}{EB6834}
\definecolor{nativelight}{HTML}{FBE3D8}
\definecolor{isol}{HTML}{2A78D6}
\definecolor{isollight}{HTML}{E3EEFB}
\definecolor{shared}{HTML}{52514E}
\definecolor{ink}{HTML}{0B0B0B}
\definecolor{ink2}{HTML}{52514E}
\definecolor{muted}{HTML}{898781}
\definecolor{faint}{HTML}{C9C7BF}
\definecolor{gridc}{HTML}{E1E0D9}
\definecolor{axisc}{HTML}{A9A79F}
\definecolor{wash}{HTML}{F4F3EF}

\pgfplotsset{
  paper/.style={
    axis line style={axisc, line width=0.4pt},
    every tick/.style={axisc, line width=0.4pt},
    major tick length=2pt, minor tick length=1.2pt,
    tick align=outside,
    axis x line*=bottom, axis y line*=left,
    tick label style={font=\scriptsize, text=ink2},
    every axis label/.append style={font=\scriptsize, text=ink},
    title style={font=\footnotesize\bfseries, text=ink, align=left},
    grid style={gridc, line width=0.3pt},
    legend style={font=\scriptsize, draw=none, fill=none, text=ink2},
    legend cell align=left,
    clip=false,
  },
}
\tikzset{
  boundline/.style={ink, dashed, line width=0.6pt},
  panel/.style={font=\footnotesize\bfseries, text=ink, anchor=south west},
  note/.style={font=\scriptsize, text=ink2, align=left},
  callout/.style={font=\scriptsize, text=ink, align=left, inner sep=3pt,
                  rounded corners=1.5pt, line width=0.5pt},
}

%% file: abstract.tex
Private fine-tuning of large language models increasingly pairs DP-SGD with low-rank adapters on frozen, quantized backbones, and every such stack inherits one unstated premise: frozen layers keep each record's clipped gradient a function of that record alone. We show that LLM.int8 breaks this premise. It routes outlier activation columns to a higher-precision path using a union over the whole batch, so one record decides how every other record's activations are computed, even though the weights are frozen and the quantization scales are per row. We bound the resulting sensitivity inflation, state a record-isolation contract that restores the standard bound, and define record-owned precision dispatch that meets it with the same GEMMs as batched LLM.int8, so isolation needs no extra arithmetic. We then audit the native CUDA stack (bitsandbytes, PEFT, Opacus) against four matched controls. A constructed LoRA--Opacus witness reaches $15.0\times$ the add/remove bound. On pretrained OPT-125M with natural text, other records alter 20\% of record--layer precision splits at batch size 16. An exploratory re-clipping analysis finds 10 of 64 candidate neighbors above the bound once every gradient is clipped; the top pair, revalidated natively, reaches $1.065C$ while every control stays within $C$, and forcing one layer's outlier mask removes the excess. A two-million-draw audit of the released adapters yields $\varepsilon$ lower bounds ($3.25$ at $\delta=10^{-5}$) below the advertised guarantee ($4.38$). Resolving an excess this small at the release level would take about 200 times more draws, whereas checking the sensitivity contract directly exposes it in a few native runs, which makes contract checking the practical audit for private quantized training.

%% file: body.tex
\section{Introduction}
\label{sec:intro}
Fine-tuning large language models on sensitive data increasingly follows one recipe: differentially private stochastic gradient descent (DP-SGD)~\citep{abadi2016} trains low-rank adapters~\citep{hu2022,yu2022} on top of a frozen backbone that is quantized to fit in accelerator memory~\citep{dettmers2022,dettmers2023}. Each component is mature, and off-the-shelf libraries compose them; NVIDIA NeMo Safe-Synthesizer, for example, pins bitsandbytes, PEFT, Opacus and Transformers together for private training~\citep{safesynth}. The privacy accountant then certifies the composition, provided that its inputs mean what it assumes. We show that whenever such a stack batches records through LLM.int8, one of those assumptions fails.

DP-SGD clips each record's gradient to norm $C$ and adds Gaussian noise calibrated to the resulting sensitivity. The argument assumes that changing one record leaves every other record's clipped gradient unchanged. DP libraries protect this premise by rejecting layers that mix examples, such as batch normalization, and they treat the physical batch size as a pure performance knob: virtual batching splits a logical batch into memory-sized pieces without changing the privacy analysis~\citep{yousefpour2021}. Frozen quantized layers pass these checks because they own no trainable parameters. Owning no parameters, however, is not the same as computing each record independently.

LLM.int8~\citep{dettmers2022} makes the gap concrete. It quantizes activations row by row, except for \emph{outlier} feature columns, which it computes in FP16. Writing $A_{r,t,:}$ for token row $t$ of record $r$, the outlier set of a batch $D$ is
\begin{equation}
O(D)=\{\,j:\ \exists\, r\in D,\,t:\ |A_{r,t,j}|\geq\tau\,\}.
\label{eq:union}
\end{equation}
Because the union ranges over all records, adding one record can move a column of \emph{every other} record from the INT8 path to the FP16 path and change their outputs and adapter gradients, even though the weights are frozen and the quantization scales are per row (Figure~\ref{fig:mechanism}). Unlike leakage through shared quantization scales~\citep{foerster2026}, no numerical statistic is pooled across the batch; what is shared is the decision of which precision path each column takes.

\input{figures/fig_mechanism}

This paper asks three questions. How much can such routing coupling inflate the sensitivity of the clipped sum, and which implementation contract rules it out? Does the native stack actually exhibit the coupling, on constructed and on natural inputs? And does the inflation reach the released model? We make the following contributions.
\begin{itemize}[leftmargin=*,itemsep=1pt,topsep=2pt]
\item \textbf{Analysis and contract} (\S\ref{sec:influence}--\ref{sec:design}). An influence bound shows that $m$ altered unchanged contributions raise add/remove sensitivity from $C$ to $(2m{+}1)C$, a limit our constructed witness attains. A record-isolation contract restores the separable bound, and record-owned precision dispatch satisfies it with the same INT8 and FP16 GEMMs as batched LLM.int8, changing only the masks.
\item \textbf{A native audit} (\S\ref{sec:method}). We audit the pinned CUDA stack along an evidence ladder (operator, LoRA--Opacus composition, pretrained OPT-125M), validate every per-record gradient against independent autograd, trace native kernel dispatch, and pair each measurement with four controls: record isolation, threshold zero, BF16 and NF4.
\item \textbf{Findings} (\S\ref{sec:results}). (i)~The constructed composition reaches $14.998C$, essentially the worst case, while all controls stay within $C$. (ii)~On natural text, other records alter 5--20\% of OPT-125M's record--layer precision splits as the physical batch grows from 2 to 16. (iii)~Once every gradient is clipped, 10 of 64 candidate neighbors exceed the bound by 1.4--6.5\% in an exploratory analysis, and the top pair reaches $1.0653C$ in 40 fresh native cases in which every control stays within $C$. (iv)~The excess traces to one column added to one layer's union; forcing that mask removes it. (v)~A two-million-draw audit of the released adapters yields $\varepsilon$ lower bounds below the advertised guarantee; a power analysis shows that resolving the measured excess at the release level would take about 200 times more draws.
\end{itemize}

\paragraph{Scope.} Our target is the proof's premise: the sensitivity contract fails natively, while the release-level audit detects no violation of the advertised guarantee. The lesson is general: a DP-SGD proof certifies a computation, not module labels. Physical batching and data-dependent numerics are part of the privacy mechanism, so they must be made record-owned or be audited. Simple configurations (singleton physical batches, threshold zero, weight-only NF4, or BF16) restored the bound in every setting we measured.

\section{Background}
\label{sec:background}
\paragraph{Privacy unit and the audited sum.}
A record owns all of its token rows, its padding and attention-mask convention, its loss terms and any private randomness. Neighboring datasets differ by adding or removing one record (our primary relation) or by replacing one. At step $t$ we condition on a fixed transcript $H_t$ of public state and previously released private outputs; persistent private state that was never released, such as a cached routing decision, is not public. The contribution that DP-SGD actually clips is
\begin{equation}
h_r(D;H_t)=g_r(D;H_t)\,\min\{1,\,C/\|g_r(D;H_t)\|_2\},
\label{eq:clip}
\end{equation}
where $g_r(D;H_t)$ is record $r$'s adapter gradient \emph{as computed inside batch $D$}, with the zero-gradient case defined by continuity. The first audited object is the clipped sum
\begin{equation}
F(D)=\textstyle\sum_{r\in D}h_r(D;H_t),
\label{eq:sum}
\end{equation}
to which DP-SGD adds $\mathcal N(0,\sigma^2C^2I)$ before averaging and the optimizer step. If each $h_r$ depends only on record $r$, then $\|F(D)-F(D')\|_2\le C$ for add/remove neighbors and $\le 2C$ for replacement: the \emph{separable bound}. A per-example interface such as Opacus's \texttt{grad\_sample} reports $g_r$, but it does not show that $g_r$ depends only on $r$. We use full sampling ($q{=}1$) and a fixed public loss divisor throughout, so that sampler or normalization effects cannot masquerade as routing effects; a data-dependent divisor under add/remove adjacency would need its own analysis.

\paragraph{LLM.int8 mixed-precision decomposition.}
For a frozen linear layer with flattened activations $A$ (all token rows of all records in the physical batch), LLM.int8 computes the outlier set $O(D)$ of Eq.~\eqref{eq:union} with threshold $\tau$ (6 by default). Each row's quantization scale is its absolute maximum over entries below $\tau$, a purely row-local statistic. The columns in $O(D)$ are then zeroed in every row's INT8 codes and recomputed in FP16, and the two partial products are added~\citep{dettmers2022}. A record's output therefore depends on other records only through $O(D)$, which decides, for every row, which columns take which precision path.

\paragraph{Threat model.} Differential privacy is a worst-case guarantee: the sensitivity bound must hold for every pair of neighboring datasets, including pairs whose differing record is chosen by an adversary who knows the model and the other records. A single measured pair whose clipped sums are farther apart than the separable bound therefore refutes that bound for the computation and lower-bounds its true sensitivity, whereas a natural pair that stays within the bound says nothing about the worst case. LLM.int8 hands such an adversary a direct lever: a record with a large activation in a column that no other record triggers moves that column to FP16 for every co-batched record. Our constructed witness (\S\ref{sec:res-witness}) shows that this lever can push the clipped sum to the limit of Proposition~\ref{prop:influence}. We do not search for adversarial natural-text records.

\section{Cross-Record Influence}
\label{sec:influence}
\subsection{How much can influence inflate sensitivity?}
\begin{proposition}[Influence bound]
\label{prop:influence}
For replacement neighbors of equal size, suppose at most $m$ unchanged contributions $h_r$ differ between $D$ and $D'$. Then
\begin{equation}
\|F(D)-F(D')\|_2\leq 2(m+1)C.
\label{eq:replace}
\end{equation}
For add/remove neighbors, if at most $m$ common contributions differ, the bound is $(2m+1)C$.
\end{proposition}
\begin{proof}
Each changed common contribution moves by at most $2C$ by the triangle inequality. Replacement adds one more pair of clipped contributions, also bounded by $2C$; add/remove adds one unmatched vector of norm at most $C$. Summing gives the bounds.
\end{proof}
Proposition~\ref{prop:influence} is elementary, and it exposes a missing premise rather than a new privacy theorem. A changed routing mask may leave gradients unchanged, changed gradients may cancel, and a pair's distance may stay below the separable bound. Conversely, a measured distance above the separable bound is a \emph{sensitivity witness}: it shows that the standard proof does not apply to that computation. The bound is tight; our constructed composition attains $(2m{+}1)C$ for $m{=}7$ (\S\ref{sec:res-witness}). A witness alone does not refute an advertised guarantee, which concerns the released output.

\subsection{From a sensitivity witness to the released output}
For the declared real-valued one-step mechanism $M(D)=F(D)+Z$ with $Z\sim\mathcal N(0,\sigma^2C^2I)$, fix a neighboring pair, let $\Delta=\|F(D)-F(D')\|_2$ and $\mu=\Delta/(\sigma C)$. Projecting the Gaussian likelihood ratio onto the mean difference gives the exact hockey-stick divergence~\citep{balle2018}
\begin{equation}
\delta_{\rm pair}(\varepsilon)=
\Phi\!\left(\frac{\mu}{2}-\frac{\varepsilon}{\mu}\right)
-e^\varepsilon\,\Phi\!\left(-\frac{\mu}{2}-\frac{\varepsilon}{\mu}\right),
\label{eq:gaussian}
\end{equation}
which is zero for $\Delta{=}0$. If $\delta_{\rm pair}$ exceeds the advertised $\delta$ at the advertised $\varepsilon$, the pair refutes the guarantee for this mechanism. A modest excess changes $\varepsilon$ modestly: $\mu{=}1.065$ instead of $1$ raises the one-step $\varepsilon$ at $\delta{=}10^{-5}$ from $4.38$ to $4.71$ (\S\ref{sec:res-release}). The test transfers to the released adapter when the output is a known invertible affine map of the noisy gradient, as for one SGD step from a fixed initialization; general optimizers or later training may lose information, and finite-precision noise is a different mechanism that must be audited rather than assumed to follow Eq.~\eqref{eq:gaussian}~\citep{mironov2012}.

\section{Record-Owned Precision Dispatch}
\label{sec:design}
\subsection{The isolation contract}
\begin{proposition}[Compositional record isolation]
\label{prop:isolation}
Condition on $H_t$. Suppose that every forward and backward operation producing record $r$'s adapter contribution reads only $r$'s data, public state and $r$'s assigned randomness; that routing decisions and persistent scratch state obey the same rule; and that numerical choices affecting the result do not depend on other records. Then each contribution is record-isolated, and Eq.~\eqref{eq:sum} has sensitivity at most $C$ for add/remove and $2C$ for replacement neighbors.
\end{proposition}
\begin{proof}
Topologically order the forward graph and the reverse-mode computations. Every record-owned input of a common record is unchanged. Each subsequent operation has unchanged admissible inputs and numerical choices, so its output is unchanged; the same induction covers backward intermediates and the final contribution. Clipping is record-local, so Proposition~\ref{prop:influence} with $m{=}0$ gives the bounds. For randomized local computation the argument holds under a fixed record-owned random tape, and a bound that holds for every tape survives averaging over tapes.
\end{proof}
The contract is sufficient, not necessary: a coupled program could admit a different valid sensitivity argument. It also does not follow that a standard GEMM backend satisfies the numerical-invariance condition, since grouping can change reductions, tile selection or backward kernels. Bitwise agreement with isolated execution is therefore a separate implementation target; if it fails, the implementation must constrain the backend or declare a different record-local numerical reference. Small numerical differences are not a noninterference proof.

\subsection{Record-owned routing semantics}
Define each record's own outlier set $O_r=\{j:\exists t,\ |A_{r,t,j}|\geq\tau\}$. The reference semantics executes LLM.int8 on all tokens of one record at a time (token-local routing is another valid choice, with a different numerical reference). A batched implementation masks each row with its owner's $O_r$, evaluates one dense integer product, and adds only that record's correction:
\begin{equation}
Y_{r,t}=Q_{\neg O_r}(A_{r,t})\,Q(W)^\top
       +A_{r,t,O_r}\,W_{:,O_r}^{\top}.
\label{eq:grouped}
\end{equation}
Here $Q_{\neg O_r}$ includes the reference scaling and dequantization; Eq.~\eqref{eq:grouped} specifies ownership, not a fused instruction sequence. Because frozen weights are public, the union of weight columns $W_{:,\cup_r O_r}$ may be fetched once per batch; the union must never decide which correction another record receives. Shared allocation, scheduling and caching are admissible only when they change no record's arithmetic, random stream or later private state.

\begin{algorithm}[t]
\caption{Record-owned LLM.int8 forward for one frozen layer. Batched LLM.int8 is the special case in which lines~5--6 use $U$ for every row.}
\label{alg:owned}
\begin{algorithmic}[1]
\REQUIRE activations $A\in\mathbb R^{n\times K}$ with row owners $\rho(i)$; INT8 weights $W_q$, scales $s_W$, dequantized $\hat W$; threshold $\tau$
\STATE $M_{ij}\leftarrow\big[\,|A_{ij}|\ge\tau\,\big]$ \COMMENT{per-entry test}
\STATE $O_r\leftarrow\{j:\exists i,\ \rho(i){=}r,\ M_{ij}\}$ for each $r$ \COMMENT{segmented}
\STATE $U\leftarrow\bigcup_r O_r$ \COMMENT{public columns to fetch}
\STATE $s_i\leftarrow\max_{j:\neg M_{ij}}|A_{ij}|$ \COMMENT{row-local scale}
\STATE $Q_{ij}\leftarrow\mathrm{round}(127A_{ij}/s_i)$ if $j\notin O_{\rho(i)}$, else $0$
\STATE $R_{ij}\leftarrow A_{ij}$ if $j\in O_{\rho(i)}$, else $0$ \quad ($j\in U$)
\STATE \textbf{return} $\mathrm{dequant}(Q W_q^{\top};s,s_W)+R\,\hat W_{:,U}^{\top}$
\end{algorithmic}
\end{algorithm}

\paragraph{Implementation.} Algorithm~\ref{alg:owned} realizes Eq.~\eqref{eq:grouped} with LLM.int8's own kernels. It needs only record offsets, which the adapter receives before flattening: a segmented reduction turns the per-entry threshold test into per-record sets $O_r$, and per-row masks replace the per-column mask. A grouped kernel can additionally skip the zeros in $R$ by issuing per-record or sparse residual products while preserving row order. The backward pass needs the same ownership audit for input-gradient products, adapter activations, clipping buffers and dropout masks: frozen layers pass data-dependent activations to trainable modules, so validating only directly trainable parameters, as Opacus does, cannot establish Proposition~\ref{prop:isolation}. Shared outlier caches must be record-scoped or cleared unless their privacy cost is accounted for.

\paragraph{Cost.} Algorithm~\ref{alg:owned} issues the same INT8 and FP16 GEMMs as batched LLM.int8, and building the sets $O_r$ reuses the threshold comparisons that LLM.int8 already performs. Ownership does not add FP16 work either: each row multiplies only its own outlier columns, so a grouped kernel that skips the zeros in $R$ does 1.5--7.5\% less correction work than batched LLM.int8 on our OPT-125M traces at $B{=}2$--$16$ (weighted by layer width; Figure~\ref{fig:coupling}a shows the unweighted share). In those traces the correction covers at most 8 of a layer's 768 or 3,072 input columns. Record ownership is therefore arithmetic-neutral by construction; the remaining engineering lies in mask construction, kernel design and numerical invariance, a clear target for an optimized kernel.

\paragraph{Simpler alternatives.} The same contract is met by singleton physical batches, threshold zero (no outlier path), a public calibrated outlier mask~\citep{ashkboos2023}, per-vector sparse outliers~\citep{hooper2024}, weight-only NF4~\citep{dettmers2023}, or an unquantized BF16 backbone. A grouped kernel is worthwhile only if it improves the resulting quality/resource frontier; native batched execution can enter an equal-privacy comparison only with a valid worst-case influence bound, never with a sampled maximum treated as global.

\section{Audit Methodology}
\label{sec:method}
\paragraph{Stack.} All experiments run on one NVIDIA H100 NVL with PyTorch 2.8.0+cu126, bitsandbytes 0.49.1~\citep{bnb}, PEFT 0.18.1~\citep{peft}, Opacus 1.5.4~\citep{opacus} and Transformers 5.12.0~\citep{transformers}, the versions pinned by Safe-Synthesizer commit \texttt{7f3b767c}~\citep{safesynth}; the operator experiment uses bitsandbytes 0.48.1. We execute the pinned components, not Safe-Synthesizer's entry point.

\begin{algorithm}[t]
\caption{The native sensitivity audit applied to OPT-125M (\S\ref{sec:res-opt}--\ref{sec:res-release}).}
\label{alg:audit}
\begin{algorithmic}[1]
\REQUIRE pinned stack and model; baseline batch $D$; candidate records $X$; clipping radii $\mathcal C$
\STATE run $D$ and every $D\cup\{x\}$, $x\in X$, natively at a screening radius
\STATE run the baseline, the top candidate and an identical repeat in each control arm
\STATE check every case: \texttt{grad\_sample} vs.\ autograd, kernel calls, clipping, zero repeat distance
\STATE re-clip the saved $g_r$ to get $\kappa(x,C)$ for all $x\in X$, $C\in\mathcal C$ \COMMENT{exploratory}
\STATE freeze the top pair; re-run it in every arm \COMMENT{validation}
\STATE diff the per-layer unions; force the changed mask in forward, then backward
\STATE replay the pair's clipped sums through native noise and update; report $\varepsilon_{\rm lb}$
\end{algorithmic}
\end{algorithm}

\paragraph{Procedure.} Algorithm~\ref{alg:audit} summarizes the audit we apply to the pretrained model: a native screen, an exploratory re-clipping analysis, native validation of the top pair, causal localization, and a release audit. The rest of this section details its arms, checks and measurements.

\paragraph{Arms.} Each experiment compares native batched execution with matched controls on identical inputs and, for adapters, identical canonical FP32 initializations: \emph{native batched INT8} (LLM.int8, $\tau{=}6$, all records in one physical batch); \emph{record-isolated INT8} (one record per physical forward, accumulated with Opacus virtual batching into a single logical noise/update step, never eight separately noised steps); \emph{threshold zero} (no outlier path); a \emph{BF16} backbone; and \emph{NF4} weight-only quantization. The four controls remove the batch-wide union in four different ways.

\paragraph{Validation.} Wherever applicable, every case checks that (i)~Opacus \texttt{grad\_sample} equals independent per-record autograd of the same realized batch loss, its sum equals the batch backward, and every trainable coordinate is covered; (ii)~each INT8 forward calls the loaded library's \texttt{cint8\_vector\_quant}, \texttt{cigemmlt\_32} and \texttt{cdequant\_mm\_int32\_fp16} entry points, and backward does not re-run quantization; (iii)~clipping matches Opacus's formula, including its $10^{-6}$ stabilizer; and (iv)~identical-batch repeats give exactly zero distance, so the measurement floor is zero. These checks separate a validated native computation from a sensitivity calculation on an emulator.

\paragraph{Measurement.} We compute $\Delta=\|F(D)-F(D')\|_2$ in FP64 from the saved FP32 contributions and report the ratio $\kappa=\Delta/C$ for add/remove and $\Delta/(2C)$ for replacement. Losses are summed with a fixed public divisor, $q{=}1$, and the noise multiplier is 1. Each case also executes one logical Opacus noise, accountant and SGD step, which verifies the release path but cannot estimate a release distribution (\S\ref{sec:res-release}). For OPT-125M, exceedance is declared only above the bound plus an absolute margin of $10^{-4}$ (plus ten times the repeat distance, which is zero), fixed in the protocol before the screen. All attempts are retained, including failed ones (Appendix~\ref{app:provenance}).

\section{Results}
\label{sec:results}
Every sensitivity rung of the evidence ladder shows the same pattern: native batching breaks the separable bound and every control stays within it. Table~\ref{tab:ladder} summarizes the ladder, from a constructed $15\times$ excess to a single column in one layer of pretrained OPT-125M; the subsections follow the same order.

\begin{table*}[t]
\caption{Evidence at a glance. Ratios are clipped-sum distances over the separable bound ($C$ for add/remove, $2C$ for replacement). Sensitivity measurements pair native batching with record-isolation, threshold-zero, BF16 and (where applicable) NF4 controls on identical inputs.}
\label{tab:ladder}
\centering\small
\setlength{\tabcolsep}{4pt}
\begin{tabularx}{\textwidth}{@{}l l l l Y@{}}
\toprule
Rung & Setting & Native result & Controls & Establishes (and does not)\\
\midrule
Operator (\S\ref{sec:res-witness}) & constructed, one INT8 layer & $7.0\times$ (replacement) & $0$ & the CUDA kernels execute the coupling (not: a natural task)\\
LoRA--Opacus (\S\ref{sec:res-witness}) & constructed, pinned stack & $15.0\times$ (add/remove) & $\le1$ & Proposition~\ref{prop:influence}'s worst case through real Opacus (not: a natural task)\\
Coupling (\S\ref{sec:res-coupling}) & OPT-125M, TinyStories, $B{=}2$--$16$ & 5--20\% of pairs coupled & -- & routing coupling is routine on natural text (not: sensitivity)\\
Sensitivity (\S\ref{sec:res-opt}) & OPT-125M, 64 candidates, 40 cases & $1.0653\times$ (add/remove) & $\le1$ & the separable premise fails natively (not: prevalence)\\
Mechanism (\S\ref{sec:res-mechanism}) & forward factorial, forced mask & one column, one layer & -- & the union change causes the excess (not: exact independence)\\
Release (\S\ref{sec:res-release}) & $2\times10^6$ actual releases & $\varepsilon_{\rm lb}{=}3.25$ vs.\ $4.38$ & -- & no detected guarantee violation (not: a privacy proof)\\
\bottomrule
\end{tabularx}
\end{table*}

\subsection{Constructed witnesses}
\label{sec:res-witness}
We first ask whether the native kernels execute the coupling at all, using a construction whose effect is easy to read (details in Appendix~\ref{app:construction}). Eight records each contribute one activation row to a frozen INT8 layer with weight $[1,0]$; two exact zeros pad the inner dimension to $K{=}4$, avoiding a narrow-$K$ floating-point fallback. An FP32 scalar bias is the only trainable parameter, $\tau{=}6$, and $C{=}0.005$. The seven unchanged rows are $[1.234,5]$; the eighth changes from $[1,5]$ to $[6.1,5]$ and crosses $\tau$ in column~0. Row scales stay at 5, but column~0 joins the union, so the unchanged rows' first entry moves from the INT8 path (output $1.220703125$) to FP16 ($1.234375$). Their common label is the midpoint of these two native outputs, fixed by a pilot, so each bias gradient flips sign ($\mp0.0068359375$) and each clipped contribution moves by $2C$. This loss geometry is chosen to make the effect legible; it is not a frequency estimate.

\begin{table}[t]
\caption{Constructed witnesses on the native CUDA stack. Ratios are clipped-sum distances over the separable bound; the last row is the worst case of Proposition~\ref{prop:influence} for $m{=}7$ altered records.}
\label{tab:witness}
\centering\small
\setlength{\tabcolsep}{4pt}
\begin{tabular}{@{}lrrr@{}}
\toprule
& Operator & \multicolumn{2}{c}{LoRA--Opacus}\\
\cmidrule(lr){2-2}\cmidrule(l){3-4}
Execution & $\Delta/2C$ & $\Delta/C$ & $\Delta/2C$\\
& \footnotesize replace & \footnotesize add/remove & \footnotesize replace\\
\midrule
Native batched INT8 & $\mathbf{7.000}$ & $\mathbf{14.998}$ & $\mathbf{6.999}$\\
Record-isolated INT8 & 0 & 1.000 & 0\\
Threshold zero & 0 & 1.000 & 0\\
BF16 & 0 & 1.000 & 0\\
NF4 & -- & 1.000 & 0\\
Identical inputs & 0 & 0 & 0\\
\midrule
Prop.~\ref{prop:influence} limit ($m{=}7$) & 8 & 15 & 8\\
\bottomrule
\end{tabular}
\end{table}

\paragraph{Operator.} The trace records the actual CUDA quantization codes, row scales, outlier indices and per-record bias derivatives. The outlier set changes from empty to $\{0\}$, and the replacement distance is exactly $14C_f$, where $C_f$ is the FP32 representation of $C$: seven times the separable bound (Table~\ref{tab:witness}). A NumPy reference of the same arithmetic gives identical outputs. Record isolation, threshold zero, BF16 and an identical-input null all give zero distance.

\paragraph{Library composition.} We replace the bias with an actual rank-one PEFT LoRA adapter (four trainable entries in $A$, one in $B$; at the $B{=}0$ initialization only $B$'s gradient is nonzero) and run Opacus per-record gradients, clipping, noise and SGD on the pinned 0.49.1 stack. Adding the eighth record changes the union from empty to $\{0\}$ and reverses all seven common records' LoRA gradients. The add/remove distance is $14.998C$, within $0.02\%$ of the $(2m{+}1)C=15C$ worst case of Proposition~\ref{prop:influence}; replacement gives $6.999\times2C$. Every control stays within its separable bound: add/remove distances equal the added record's own contribution ($0.99999979C$), and replacement distances are zero. In all 20 measurements (5 arms $\times$ 4 inputs) Opacus \texttt{grad\_sample} matched independent autograd on the same realized batch. Under the declared real-valued Gaussian mechanism (Eq.~\eqref{eq:gaussian}), a pair this far apart would raise the one-step $\varepsilon$ at $\delta{=}10^{-5}$ and $\sigma{=}1$ from 4.38 to about 176; we did not audit the finite-precision release of this constructed pair.

\subsection{Coupling is routine on natural text}
\label{sec:res-coupling}
Does the pathway carry traffic on natural inputs? We instrument pretrained OPT-125M~\citep{zhang2022opt} with native LLM.int8 ($\tau{=}6$) and rank-8 LoRA adapters on the query and value projections (294,912 trainable parameters in 48 tensors), on the first 16 TinyStories training windows~\citep{eldan2023} (256 tokens each, official padding masks, dropout off, masked next-token loss with fixed public divisor 255). At physical batch sizes 2, 4, 8 and 16, every configuration passes Opacus validation and executes a real clipping, noise and optimizer step with complete per-record gradients. Each step issues 72 native INT8 quantization, integer-GEMM and dequantization calls; 42 of the 72 layers have a nonempty outlier union, which matches the union recomputed from the $B{\times}256$ activation rows exactly.

\input{figures/fig_coupling}

For every layer we also record each record's own set $O_r$. Figure~\ref{fig:coupling}a shows how often it differs from the union. At $B{=}2$, 4.8\% of (record, layer) pairs in the 42 outlier-carrying layers compute some column in FP16 only because another record crossed $\tau$; at $B{=}16$ the share is 20.1\%, and 7.0\% of all FP16 columns applied to a record were triggered by others. Figure~\ref{fig:coupling}b shows where this happens at $B{=}16$: every record is coupled in layer~7's attention input, as are 13 and 14 records in layer~9's attention input and layer~11's first feed-forward projection, while 15 of the 22 outlier-carrying cells show no coupling at all. Most outlier columns are shared by every record, consistent with the systematic outlier features documented by \citet{dettmers2022}. Only 3--7 columns per batch are owned by a single record, and only those can leave the union when one record is removed. Whether such a change moves a clipped gradient is the question of the next experiment.

\subsection{Sensitivity on pretrained OPT-125M}
\label{sec:res-opt}
We fix a seven-record baseline $D$ (TinyStories rows 16--22) and consider adding each of 64 candidate records (rows 24--87), keeping every outcome. The initial screen at $C{=}1$ is negative: over the 64 candidates, the largest native add/remove distance is $0.99999921C$.

\input{figures/fig_landscape}

\paragraph{Why the screen missed.} Re-clipping the saved native per-record gradients at nine radii (Figure~\ref{fig:landscape}a; 576 exploratory scores, none discarded) explains the negative result. At $C{=}1$ most per-record gradients are shorter than $C$, so the added record contributes less than $C$ and absorbs the cross-record change. Once every gradient is clipped ($C\le0.25$), $\kappa$ no longer depends on $C$, and 10 of the 64 candidates exceed the bound by 1.4--6.5\%. Small radii at which most gradients are clipped are common in private language-model fine-tuning~\citep{li2022}, so this regime is practically relevant.

\paragraph{Native validation.} We then froze the top-ranked candidate (row 75) and ran 40 fresh native cases: the baseline, an earlier addition (row 23), the selected addition and an identical repeat, under native batching and all four controls, at $C{=}0.001$ (the top-ranked radius) and $C{=}0.25$. All 40 cases pass the execution, initialization, derivative and clipping checks. Native batching gives $\Delta=0.2663144763=1.065257905C$ at $C{=}0.25$, above the bound plus the protocol margin. At $C{=}0.001$ it gives the same ratio, $1.065257904$, but the absolute excess ($6.5\times10^{-5}$) is below the $10^{-4}$ margin; since $\kappa$ is scale-free in this regime the two runs measure the same relative excess, and we report the margin outcome as declared. The excess is four orders of magnitude larger than any control's deviation from $C$ (at most $2.4\times10^{-6}$ for isolation, threshold zero and BF16, and $4.7\times10^{-4}$ below $C$ for NF4; Figure~\ref{fig:landscape}b). All replacement ratios, including the native $0.671$, stay below $2C$. Because sensitivity is a supremum over neighboring datasets, this one pair already shows that the native computation's add/remove sensitivity is at least $1.0653C$, not $C$. Selection and validation belong to one discovery study on one baseline batch, so these results show that the separable premise fails for a natural pretrained configuration; they do not estimate how often it fails across batches or models.

\subsection{One layer's mask causes the excess}
\label{sec:res-mechanism}
Across all 72 INT8 layers, exactly one union differs between $D$ and $D'$ (Figure~\ref{fig:intervention}a): in decoder layer~6's first feed-forward projection, adding row 75 grows the union from $\{422\}$ to $\{422,706\}$. The seven common records' LoRA gradients differ by 0.048--0.071 in $\ell_2$ norm.

\input{figures/fig_intervention}

\paragraph{Forward factorial.} We capture the common records' input activations at that layer under $D$ and $D'$ and cross the two value tensors with the two masks, with the input batch fixed to $D$, the original quantization kernel and FP16 correction, and the other 71 layers unmodified; an intervention is rejected if its forced union would omit a natural outlier. All seven forwards (two native references, an identical repeat and four factorial cells) complete. Reference losses and all 72 masks match the parent experiment bitwise, and both diagonal cells reproduce their native logits bitwise. The captured activations are bitwise identical under $D$ and $D'$, so changing only the values has no effect, whereas changing only the mask reproduces the entire common-record logit change (maximum absolute difference $0.36630$, $\ell_2$ difference $309.3763$; Figure~\ref{fig:intervention}b).

\paragraph{Backward intervention.} Four further native Opacus cases force that layer to use the frozen union $O(D')$ in both batches, keeping the original inputs, autograd function and backward path; the two unforced cases reproduce all six saved parent arrays bitwise, and an independent CPU reconciliation checks every saved array. Under the forced union the clipped-sum distance falls from $1.065257905C$ to $0.999997838C$, and each common record's gradient change shrinks from 0.048--0.071 to $1.1$--$2.7\times10^{-6}$ (Figure~\ref{fig:intervention}c). The residual is not zero (largest coordinate difference $1.1\times10^{-7}$), so exact independence is not established; we have not attributed the residual, and batch-shape-dependent kernel numerics, which Proposition~\ref{prop:isolation} also excludes, are one candidate. Because the forced mask is chosen post hoc from $D'$, the intervention localizes the cause; it is not a deployable private policy.

\subsection{Does the excess reach the released model?}
\label{sec:res-release}
Sensitivity describes an intermediate sum; a privacy guarantee describes the released output. We therefore audit actual releases of the selected pair at $C{=}0.25$. The audit replays the accepted deterministic clipped sums for $D$ and $D'$ through unmodified Opacus 1.5.4 noising and scaling and Torch SGD (learning rate 0.01, noise multiplier 1, divisor 1), resetting all 294,912 FP32 adapter coordinates before each draw. Anchors reproduce both parent noised gradients and released adapters bitwise. Forward, backward and clipping are not re-run, and noise comes from the pinned non-cryptographic CUDA generator.

\input{figures/fig_release}

Eight 250,000-draw shards give $10^6$ releases per neighbor, two million in total, excluding qualification and warm-up draws. Each shard takes 12--13 minutes ($\approx$3\,ms per release; 1.7 GPU-hours overall). The test statistic projects the released FP32 adapter displacement, in FP64, onto the clipped-sum difference direction, scaled by the learning rate times $\sigma C$; internal gradients are used only as diagnostics. For each $\delta\in\{10^{-3},10^{-5},10^{-6}\}$, half of each neighbor's draws select a threshold among 1,001 pooled quantiles by maximizing the empirical log-ratio objective (ties go to the lowest threshold), and the other half evaluates it with a strict inequality, without re-selection. One-sided Clopper--Pearson bounds at $\alpha=0.05/6$ (nominal family-wise 95\% over the three $\delta$) give a lower bound $p_L$ on the $D'$ tail and an upper bound $q_U$ on the $D$ tail, and
\begin{equation}
\varepsilon_{\rm lb}=\max\{0,\ \log((p_L-\delta)/q_U)\}.
\end{equation}

All three $\delta$ select the same threshold, $5.2083$. Held-out counts are $1{,}043/500{,}000$ for $D'$ and $23/500{,}000$ for $D$, so $p_L=1.935\times10^{-3}$ and $q_U=7.456\times10^{-5}$. The resulting bounds (2.53, 3.25 and 3.26 at $\delta=10^{-3}$, $10^{-5}$ and $10^{-6}$) lie below the ideal real-Gaussian curves both at the advertised $\mu{=}1$ (3.14, 4.38, 4.89) and at the measured $\mu{=}1.065$ (3.39, 4.71, 5.25; Figure~\ref{fig:release}a). The audit detects no violation of the advertised guarantee.

\paragraph{How large would an audit need to be?} The audit is resolution-limited. Plugging the expected tail counts of an \emph{ideal} Gaussian release into the same Clopper--Pearson procedure, even with an oracle threshold, yields only about 3.1 ($\mu{=}1$) and 3.3 ($\mu{=}1.065$) at $\delta{=}10^{-5}$ with $5\times10^5$ held-out draws per neighbor; the observed 3.25 lies between them, so the finite-precision release is consistent with an ideal Gaussian at the measured sensitivity. Repeating the calculation for larger audits (Figure~\ref{fig:release}b), the expected bound for the measured pair first exceeds the advertised $\varepsilon{=}4.38$ at about $1.0\times10^8$ held-out draws per neighbor: 206 times our audit, or roughly 350 GPU-hours at our replay cost. The native sensitivity check of \S\ref{sec:res-opt} exposed the same excess deterministically, in a handful of cases. For small excesses, verifying the sensitivity contract directly is far cheaper than detecting its consequence in the release. Nominal Monte Carlo coverage also assumes that the pinned generator represents the intended distribution, so the audit is not a formal finite-precision privacy proof.

\section{Related Work}
\label{sec:related}
\paragraph{Private fine-tuning.} DP-SGD~\citep{abadi2016} with parameter-efficient adapters is a standard route to private language-model fine-tuning~\citep{yu2022,li2022}. Opacus~\citep{yousefpour2021} computes per-example gradients and its validator rejects known cross-example layers such as batch normalization; the JAX Privacy documentation explicitly warns about cross-example operations and recommends per-example execution where possible~\citep{jaxprivacy}. Record isolation is thus an established obligation, not our invention; we show that a frozen, parameter-free quantized layer violates it in a widely used stack and measure the consequences.

\paragraph{Implementation-level DP failures.} Floating-point implementations can break mechanisms that are private over the reals~\citep{mironov2012}, DP libraries widely underestimate sensitivity because of finite-precision arithmetic~\citep{casacuberta2022}, and privacy auditing has exposed bugs in DP training code~\citep{tramer2022}. The failure we study is of a different kind: one record's arithmetic depends on other records through a data-dependent precision routing decision.

\paragraph{Privacy auditing.} Auditing DP-SGD with instantiated neighbors and Clopper--Pearson lower bounds on $\varepsilon$ is standard~\citep{jagielski2020,nasr2021}, and tight or single-run variants reduce its cost~\citep{nasr2023,steinke2023}. We use a fixed-pair audit of actual released adapters to test whether a measured sensitivity excess reaches the output; our result shows how many draws such an audit needs before it can resolve a small excess.

\paragraph{Cross-example flows in batched execution.} Batch normalization couples examples and is excluded by DP libraries. Quantamination shows that dynamic quantization with shared scales leaks data across a batch~\citep{foerster2026}, and Architectural Backdoors develops within-batch data stealing together with batch noninterference checks that include dynamic quantization~\citep{kuchler2026}. We build on this boundary. In LLM.int8 the scales are row-local and the coupling runs through outlier-column ownership; we measure its effect on clipped DP-SGD sensitivity and on the released update.

\paragraph{Quantized training and inference.} LLM.int8~\citep{dettmers2022} and QLoRA~\citep{dettmers2023} enable fine-tuning on quantized backbones. QUIK selects outlier columns by public calibration~\citep{ashkboos2023} and KVQuant keeps sparse outliers per vector~\citep{hooper2024}; both avoid batch-dependent routing, and their optimized kernels are the baselines any grouped record-owned kernel must beat. DPQuant studies how quantization affects the utility of private training~\citep{gao2026}; it does not address the sensitivity premise studied here.

\section{Discussion}
\label{sec:discussion}
\paragraph{Implications for system builders.} First, validate data dependence, not parameter ownership. Opacus's validator accepts OPT-125M at every batch size we tried, although batching exceeds the bound; a batch-invariance test that compares each record's \texttt{grad\_sample} under native batching with singleton execution, as our isolation control does, would have flagged it. Second, treat the physical batch size as privacy-relevant unless per-record computation is shown to be batch-independent: virtual batching is privacy-neutral only under that condition, and an expected Poisson batch of one does not by itself guarantee singleton execution. Third, derive routing from public calibration or from already released state; reading the current private batch before a valid release is a different operation. Fourth, record a runtime manifest read from the executing objects (package versions, kernel dispatch, trainable flags, physical and logical batch sizes, loss normalization, clipping, sampler, accountant and noise settings), so that an auditor can check the computation the accountant certifies rather than the configuration file. Record-local configurations restored the bound in all our experiments, but they have different costs: threshold zero removes the outlier path whose absence degrades large models~\citep{dettmers2022}, singleton execution costs throughput, and BF16 costs memory.

\paragraph{Beyond LLM.int8.} Any decision computed over the physical batch is a candidate channel of the same kind: activation scales computed per tensor rather than per row, as in dynamically scaled FP8 or INT8 recipes~\citep{micikevicius2022,foerster2026}; capacity-limited mixture-of-experts routing, which drops a token when its expert overflows~\citep{fedus2022}; and kernels whose reduction order depends on the batch shape. None of these needs a trainable parameter to violate Proposition~\ref{prop:isolation}, and each can be screened with the same batch-invariance test.

\paragraph{Scope and future work.} Our natural-data evidence covers one model (OPT-125M), one corpus (TinyStories) and one baseline batch, with discovery, radius selection, validation and auditing in one exploratory study; measuring prevalence is the next step. Larger OPT models, where outlier features are more pronounced~\citep{dettmers2022}, an independently pretrained lineage and a second corpus are natural targets. We analyze one step with full sampling; multi-step training under Poisson sampling and its accounting are not audited. We executed Safe-Synthesizer's pinned components but not its entry point, whose default physical batch of one is a null case for the union. Our controls and interventions do not verify every backward dispatch, schedule or cache. Finally, record-owned dispatch is specified but not yet implemented as an optimized kernel. A fair evaluation would compare it with optimized singleton or vectorized isolation, a public calibrated mask with fused correction, per-vector sparse residuals, BF16 and NF4 at equal privacy, with the checkpoint, privacy unit, adjacency, logical batch, sampler, loss, $C$, adapter rank, data order and tuning budget held fixed.

\section{Conclusion}
\label{sec:conclusion}
Frozen weights do not make records independent. In LLM.int8, a batch-wide outlier union lets one record set the precision of every other record's activations, and the native bitsandbytes, PEFT and Opacus stack executes this path. A constructed composition reaches $15\times$ the DP-SGD add/remove bound, and on pretrained OPT-125M with natural text a selected neighbor exceeds it by 6.5\% once clipping is active, an excess traced to one column in one layer. A two-million-draw audit finds no release-level violation; what fails is the proof's premise. Detecting an excess this small in released outputs would take about 200 times more draws, whereas checking the sensitivity contract exposes it in a few native runs. Because record-owned routing needs no extra arithmetic and simple record-local configurations restore the bound, private training stacks should enforce, or test, record ownership of data-dependent numerics instead of inferring it from frozen parameters.

\section*{AI Usage Disclosure}
We used ChatGPT (OpenAI) to refine the manuscript's language and formatting, and Claude (Anthropic) to assist with figure preparation. AI assistance was limited to these presentation-related tasks and did not extend to research ideation, literature review, experimental design, research code generation, or the development of technical content. We reviewed and approved all AI-assisted edits and figures and take full responsibility for the final manuscript.

%% file: figures/fig_mechanism.tex
\newcommand{\figcell}[3]{
  \fill[#3] ($(#1,#2)+(0.02,-0.02)$) rectangle ($(#1,#2)+(\figs-0.02,-\figs+0.02)$)}
\newcommand{\figmatrix}{%
  \foreach \i/\tint in {0/wash,1/wash,2/wash,3/wash,4/nativelight,5/nativelight}{%
    \pgfmathsetmacro{\yi}{-(\i*\figs + int(\i/2)*\figg)}%
    \foreach \j in {0,...,7}{\figcell{\j*\figs}{\yi}{\tint};}}%
  \foreach \i/\j/\c in {0/2/shared,3/2/shared,4/2/shared,5/5/native}{%
    \pgfmathsetmacro{\yi}{-(\i*\figs + int(\i/2)*\figg)}%
    \figcell{\j*\figs}{\yi}{\c};}%
  \foreach \k/\lab in {0/{$r_1$},1/{$r_2$},2/{$x$}}{%
    \node[font=\footnotesize, text=ink, anchor=east] at (-0.04,{-(2*\k*\figs+\k*\figg)-\figs}) {\lab};}%
}
\newcommand{\figmask}[3]{
  \pgfmathsetmacro{\yb}{-(2*#1*\figs + #1*\figg) - 0.5*\figs}%
  \foreach \j in {0,...,7}{%
    \draw[faint, line width=0.4pt, fill=white] ($(#2+\j*\figs,\yb)+(0.035,-0.035)$) rectangle ($(#2+\j*\figs,\yb)+(\figs-0.035,-\figs+0.035)$);}%
  \foreach \j/\c in {#3}{\figcell{#2+\j*\figs}{\yb}{\c};}%
}
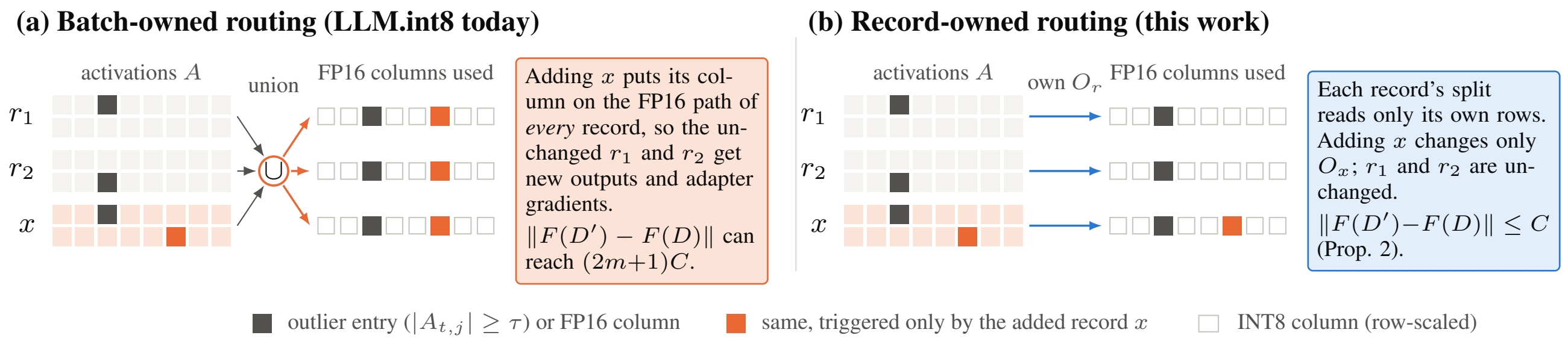
\begin{figure*}[t]
\centering
\begin{tikzpicture}[x=1cm,y=1cm,font=\scriptsize]
\def\figs{0.25}\def\figg{0.1}
\def\mx{2.9}
\def\ux{2.45}
\pgfmathsetmacro{\ymid}{-(3*\figs+\figg)}
\begin{scope}
  \node[panel] at (-0.5,0.5) {(a) Batch-owned routing (LLM.int8 today)};
  \node[note, anchor=south] at (4*\figs,0.03) {activations $A$};
  \node[note, anchor=south] at (\ux,-0.11) {union};
  \node[note, anchor=south] at (\mx+4*\figs,0.03) {FP16 columns used};
  \figmatrix
  \node[circle, draw=native, fill=white, line width=0.7pt, inner sep=0.4pt, font=\small, text=ink] (u) at (\ux,\ymid) {$\cup$};
  \foreach \k in {0,1,2}{%
    \pgfmathsetmacro{\yk}{-(2*\k*\figs+\k*\figg)-\figs}
    \draw[-{Latex[length=1.4mm]}, ink2, line width=0.4pt, shorten >=0.5pt] (2.05,\yk) -- (u);
    \draw[-{Latex[length=1.4mm]}, native, line width=0.65pt, shorten <=0.5pt] (u) -- (\mx-0.05,\yk);}
  \figmask{0}{\mx}{2/shared,5/native}
  \figmask{1}{\mx}{2/shared,5/native}
  \figmask{2}{\mx}{2/shared,5/native}
  \node[callout, draw=native, fill=nativelight, text width=2.55cm, anchor=west] at (5.1,\ymid)
    {Adding $x$ puts its column on the FP16 path of \emph{every} record, so the unchanged $r_1$ and $r_2$ get new outputs and adapter gradients.\\[2pt]
     $\|F(D')-F(D)\|$ can reach $(2m{+}1)C$.};
\end{scope}
\draw[gridc, line width=0.5pt] (8.18,0.62) -- (8.18,-1.95);
\begin{scope}[xshift=8.69cm]
  \node[panel] at (-0.5,0.5) {(b) Record-owned routing (this work)};
  \node[note, anchor=south] at (4*\figs,0.03) {activations $A$};
  \node[note, anchor=south] at (\ux,-0.11) {own $O_r$};
  \node[note, anchor=south] at (\mx+4*\figs,0.03) {FP16 columns used};
  \figmatrix
  \foreach \k in {0,1,2}{%
    \pgfmathsetmacro{\yk}{-(2*\k*\figs+\k*\figg)-\figs}
    \draw[-{Latex[length=1.4mm]}, isol, line width=0.65pt] (2.05,\yk) -- (\mx-0.05,\yk);}
  \figmask{0}{\mx}{2/shared}
  \figmask{1}{\mx}{2/shared}
  \figmask{2}{\mx}{2/shared,5/native}
  \node[callout, draw=isol, fill=isollight, text width=2.55cm, anchor=west] at (5.1,\ymid)
    {Each record's split reads only its own rows. Adding $x$ changes only $O_x$; $r_1$ and $r_2$ are unchanged.\\[2pt]
     $\|F(D')-F(D)\|\le C$ (Prop.~\ref{prop:isolation}).};
\end{scope}
\begin{scope}[yshift=-2.4cm, xshift=2.2cm]
  \figcell{0}{0}{shared}; \node[note, anchor=west] at (0.28,-0.125) {outlier entry ($|A_{t,j}|\ge\tau$) or FP16 column};
  \figcell{5.2}{0}{native}; \node[note, anchor=west] at (5.48,-0.125) {same, triggered only by the added record $x$};
  \draw[faint, line width=0.4pt, fill=white] (10.4,-0.035) rectangle (10.615,-0.215); \node[note, anchor=west] at (10.7,-0.125) {INT8 column (row-scaled)};
\end{scope}
\end{tikzpicture}
\caption{\textbf{Frozen weights do not make records independent.} LLM.int8 sends an activation column to an FP16 path if \emph{any} row of the flattened batch reaches the threshold $\tau$, and applies that batch-wide set to every row. (a)~Adding or changing one record ($x$) therefore changes the arithmetic of unchanged records ($r_1,r_2$), although all weights are frozen and quantization scales are per row. (b)~Record-owned routing derives each record's outlier set $O_r$ from its own rows, which restores the separable clipping bound.}
\label{fig:mechanism}
\end{figure*}

%% file: figures/fig_coupling.tex
\begin{figure*}[t]
\centering
\begin{tikzpicture}
\begin{axis}[paper, name=bars,
  at={(0,0)}, anchor=south west,
  width=6.1cm, height=2.45cm, scale only axis,
  ybar=1.2pt, bar width=9pt,
  symbolic x coords={2,4,8,16}, xtick=data,
  enlarge x limits=0.17,
  ymin=0, ymax=24, ytick={0,5,10,15,20}, ymajorgrids,
  yticklabel={\pgfmathprintnumber{\tick}\%},
  xlabel={Physical batch size $B$ (records per forward pass)},
  legend style={at={(0,1.04)}, anchor=south west, legend columns=1, row sep=-1pt},
  legend image code/.code={\fill[#1] (0cm,-0.07cm) rectangle (0.22cm,0.09cm);},
]
\addplot[fill=native, draw=none,
  nodes near coords={\pgfmathprintnumber[fixed,precision=1]{\pgfplotspointmeta}\%},
  nodes near coords style={font=\scriptsize, text=ink, yshift=-1pt}]
  table[x=B, y=pct_coupled_pairs, col sep=comma, row sep=crcr] {
B,int8_layers,nonempty_layers,record_layer_pairs,coupled_pairs,pct_coupled_pairs,owned_columns,foreign_columns,pct_foreign_columns,uniquely_owned_columns\\
2,72,42,84,4,4.76,330,4,1.2,4\\
4,72,42,168,24,14.29,666,30,4.31,7\\
8,72,42,336,55,16.37,1338,86,6.04,3\\
16,72,42,672,135,20.09,2650,198,6.95,3\\
};
\addplot[fill=faint, draw=none]
  table[x=B, y=pct_foreign_columns, col sep=comma, row sep=crcr] {
B,int8_layers,nonempty_layers,record_layer_pairs,coupled_pairs,pct_coupled_pairs,owned_columns,foreign_columns,pct_foreign_columns,uniquely_owned_columns\\
2,72,42,84,4,4.76,330,4,1.2,4\\
4,72,42,168,24,14.29,666,30,4.31,7\\
8,72,42,336,55,16.37,1338,86,6.04,3\\
16,72,42,672,135,20.09,2650,198,6.95,3\\
};
\legend{record--layer pairs whose FP16 set includes another record's column,
        FP16 columns applied to a record that it did not trigger}
\end{axis}
\node[panel] at ($(bars.north west)+(-0.95cm,0.78cm)$) {(a) Coupling grows with the physical batch};
\begin{scope}[xshift=9.25cm, yshift=2.45cm, x=1cm, y=1cm]
  \def\cw{0.58}\def\ch{0.5}
  \node[panel] at (-2.25,0.78) {(b) Where it happens ($B{=}16$, 72 INT8 layers)};
  \foreach \r/\lab in {0/{attn input (q, k, v)}, 1/{attn output}, 2/{fc1}, 3/{fc2}}{
    \node[note, anchor=east, text=ink] at (-0.06,{-(\r+0.5)*\ch}) {\lab};}
  \pgfplotstableread[col sep=comma, row sep=crcr] {
row,module,layer,union_size,coupled_records\\
0,attn input,0,0,0\\
0,attn input,1,0,0\\
0,attn input,2,6,0\\
0,attn input,3,6,0\\
0,attn input,4,7,5\\
0,attn input,5,5,0\\
0,attn input,6,6,2\\
0,attn input,7,8,16\\
0,attn input,8,4,0\\
0,attn input,9,3,13\\
0,attn input,10,2,0\\
0,attn input,11,2,0\\
1,attn output,0,0,0\\
1,attn output,1,0,0\\
1,attn output,2,0,0\\
1,attn output,3,0,0\\
1,attn output,4,0,0\\
1,attn output,5,0,0\\
1,attn output,6,0,0\\
1,attn output,7,0,0\\
1,attn output,8,0,0\\
1,attn output,9,0,0\\
1,attn output,10,0,0\\
1,attn output,11,0,0\\
2,fc1,0,0,0\\
2,fc1,1,4,0\\
2,fc1,2,2,0\\
2,fc1,3,1,0\\
2,fc1,4,1,0\\
2,fc1,5,1,8\\
2,fc1,6,1,0\\
2,fc1,7,1,0\\
2,fc1,8,2,0\\
2,fc1,9,2,0\\
2,fc1,10,2,5\\
2,fc1,11,6,14\\
3,fc2,0,8,0\\
3,fc2,1,0,0\\
3,fc2,2,0,0\\
3,fc2,3,0,0\\
3,fc2,4,0,0\\
3,fc2,5,0,0\\
3,fc2,6,0,0\\
3,fc2,7,0,0\\
3,fc2,8,0,0\\
3,fc2,9,0,0\\
3,fc2,10,0,0\\
3,fc2,11,0,0\\
}\couplingtab
  \pgfplotstablegetrowsof{\couplingtab}
  \pgfmathtruncatemacro{\lastrow}{\pgfplotsretval-1}
  \foreach \k in {0,...,\lastrow}{
    \pgfplotstablegetelem{\k}{row}\of\couplingtab \let\rr\pgfplotsretval
    \pgfplotstablegetelem{\k}{layer}\of\couplingtab \let\ll\pgfplotsretval
    \pgfplotstablegetelem{\k}{union_size}\of\couplingtab \let\uu\pgfplotsretval
    \pgfplotstablegetelem{\k}{coupled_records}\of\couplingtab \let\cc\pgfplotsretval
    \pgfmathsetmacro{\xa}{\ll*\cw+0.025}\pgfmathsetmacro{\xb}{(\ll+1)*\cw-0.025}
    \pgfmathsetmacro{\ya}{-\rr*\ch-0.025}\pgfmathsetmacro{\yb}{-(\rr+1)*\ch+0.025}
    \ifnum\uu=0
      \draw[gridc, line width=0.4pt] (\xa,\ya) rectangle (\xb,\yb);
    \else
      \pgfmathtruncatemacro{\pct}{12+88*\cc/16}
      \ifnum\cc=0 \def\cellfill{wash}\else\edef\cellfill{native!\pct!wash}\fi
      \fill[\cellfill] (\xa,\ya) rectangle (\xb,\yb);
      \ifnum\pct>55 \def\tc{white}\else\def\tc{ink2}\fi
      \node[font=\scriptsize, text=\tc] at ({(\ll+0.5)*\cw},{-(\rr+0.5)*\ch}) {\cc};
    \fi
  }
  \foreach \l in {0,...,11}{\node[note] at ({(\l+0.5)*\cw},{-4*\ch-0.17}) {\l};}
  \node[note, anchor=north] at ({6*\cw},{-4*\ch-0.3}) {decoder layer};
\end{scope}
\end{tikzpicture}
\caption{\textbf{Coupling is routine on natural text.} Native LLM.int8 forward passes of pretrained OPT-125M on the first 16 TinyStories training windows. (a)~Of the 72 INT8 layers, 42 have a nonempty outlier union at every batch size; the bars count (record, layer) pairs in those 42 layers. At $B{=}16$, one pair in five computes at least one column in FP16 only because another record crossed the threshold. (b)~Number of records, out of 16, whose own outlier set differs from the batch union in each layer; blank cells have no outlier columns. The query, key and value projections read the same input and share one row. Coupling concentrates in a few layers: every record is affected in layer~7's attention input, but most layers with outliers show none, because their outlier columns are shared by all records.}
\label{fig:coupling}
\end{figure*}
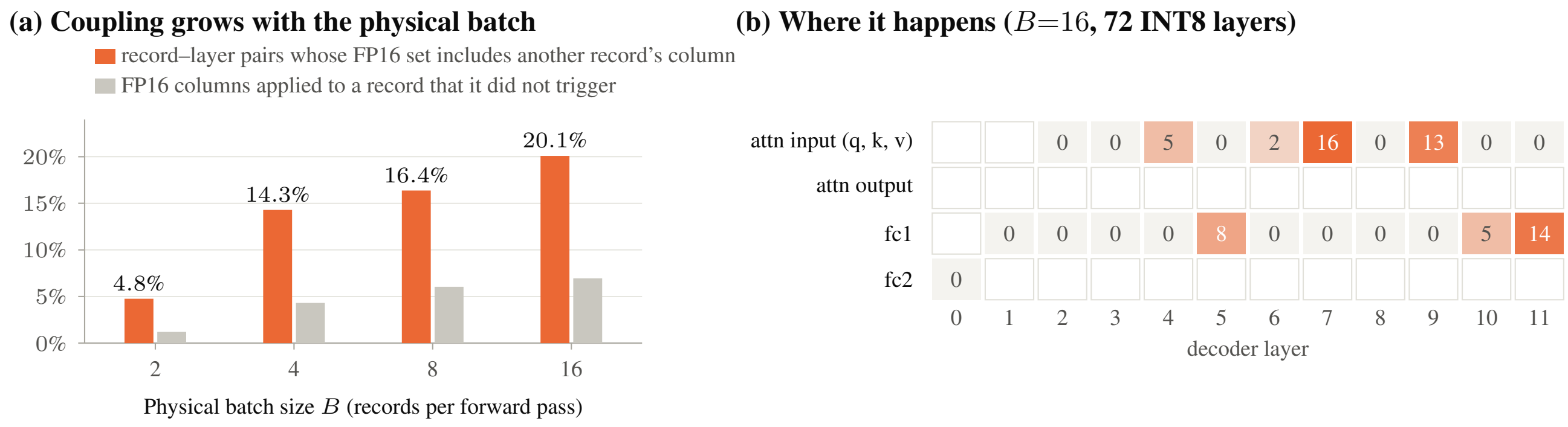

%% file: figures/fig_landscape.tex
\begin{figure*}[t]
\centering
\begin{tikzpicture}
\begin{axis}[paper, name=top, clip=true, clip mode=individual,
  width=9.4cm, height=2.3cm, scale only axis,
  xmode=log, xmin=7e-5, xmax=5.5,
  ymin=0.994, ymax=1.114, ytick={1,1.02,1.04,1.06}, ymajorgrids,
  yticklabel style={/pgf/number format/fixed, /pgf/number format/precision=2},
  axis x line=none, axis y line*=left,
]
\fill[wash] (axis cs:7e-5,0.994) rectangle (axis cs:0.25,1.114);
\draw[boundline] (axis cs:7e-5,1) -- (axis cs:5.5,1);
\addplot[faint, line width=0.45pt, unbounded coords=jump] table[x=C, y=kappa, col sep=comma, row sep=crcr] {
candidate,C,kappa\\
24,0.0001,0.9999973708636837\\
24,0.001,0.9999973708636851\\
24,0.01,0.9999973708636846\\
24,0.1,0.9999973708636841\\
24,0.25,0.9999973708636861\\
24,0.5,0.9999973445230439\\
24,1.0,0.5347295375545472\\
24,2.0,0.2673647687772736\\
24,4.0,0.1336823843886368\\
24,nan,nan\\
25,0.0001,0.9999985062968676\\
25,0.001,0.9999985062968694\\
25,0.01,0.9999985062968673\\
25,0.1,0.9999985062968682\\
25,0.25,0.9999985062968669\\
25,0.5,0.9999984681413554\\
25,1.0,0.5672308555343831\\
25,2.0,0.28361542776719156\\
25,4.0,0.14180771388359578\\
25,nan,nan\\
27,0.0001,0.999996724697214\\
27,0.001,0.9999967246972137\\
27,0.01,0.9999967246972137\\
27,0.1,0.9999967246972143\\
27,0.25,0.9999967246972133\\
27,0.5,0.9754419512151915\\
27,1.0,0.48772070114125404\\
27,2.0,0.24386035057062702\\
27,4.0,0.12193017528531351\\
27,nan,nan\\
28,0.0001,0.9999981632548344\\
28,0.001,0.9999981632548344\\
28,0.01,0.9999981632548347\\
28,0.1,0.9999981632548339\\
28,0.25,0.9999981632548343\\
28,0.5,0.999998120777037\\
28,1.0,0.6092633930875981\\
28,2.0,0.30463169654379907\\
28,4.0,0.15231584827189953\\
28,nan,nan\\
29,0.0001,0.9999976969124461\\
29,0.001,0.9999976969124469\\
29,0.01,0.9999976969124467\\
29,0.1,0.9999976969124463\\
29,0.25,0.9999976969124469\\
29,0.5,0.9999976520437359\\
29,1.0,0.6055618202276727\\
29,2.0,0.30278091011383634\\
29,4.0,0.15139045505691817\\
29,nan,nan\\
30,0.0001,0.9999975290595678\\
30,0.001,0.9999975290595662\\
30,0.01,0.9999975290595672\\
30,0.1,0.9999975290595676\\
30,0.25,0.9999975290595675\\
30,0.5,0.9999974981435457\\
30,1.0,0.6246483026938584\\
30,2.0,0.3123241513469292\\
30,4.0,0.1561620756734646\\
30,nan,nan\\
31,0.0001,0.9999981722033685\\
31,0.001,0.9999981722033685\\
31,0.01,0.9999981722033672\\
31,0.1,0.999998172203368\\
31,0.25,0.9999981722033684\\
31,0.5,0.9999981226395972\\
31,1.0,0.526259355638783\\
31,2.0,0.2631296778193915\\
31,4.0,0.13156483890969575\\
31,nan,nan\\
32,0.0001,0.999996957741305\\
32,0.001,0.999996957741305\\
32,0.01,0.9999969577413069\\
32,0.1,0.999996957741306\\
32,0.25,0.9999969577413058\\
32,0.5,0.9999969276567893\\
32,1.0,0.640132969444828\\
32,2.0,0.320066484722414\\
32,4.0,0.160033242361207\\
32,nan,nan\\
33,0.0001,1.0000000461266045\\
33,0.001,1.0000000461266056\\
33,0.01,1.0000000461266048\\
33,0.1,1.0000000461266043\\
33,0.25,1.0000000461266036\\
33,0.5,0.9999999821715908\\
33,1.0,0.5599606374668892\\
33,2.0,0.2799803187334446\\
33,4.0,0.1399901593667223\\
33,nan,nan\\
34,0.0001,0.9999984200634098\\
34,0.001,0.9999984200634093\\
34,0.01,0.9999984200634099\\
34,0.1,0.999998420063411\\
34,0.25,0.9999984200634093\\
34,0.5,0.9999983790130175\\
34,1.0,0.5599851075788533\\
34,2.0,0.27999255378942667\\
34,4.0,0.13999627689471333\\
34,nan,nan\\
36,0.0001,0.999998479259472\\
36,0.001,0.9999984792594708\\
36,0.01,0.9999984792594717\\
36,0.1,0.9999984792594718\\
36,0.25,0.9999984792594707\\
36,0.5,0.9999984285593985\\
36,1.0,0.581410361047393\\
36,2.0,0.2907051805236965\\
36,4.0,0.14535259026184824\\
36,nan,nan\\
37,0.0001,0.9999979102196204\\
37,0.001,0.9999979102196198\\
37,0.01,0.9999979102196191\\
37,0.1,0.999997910219619\\
37,0.25,0.9999979102196198\\
37,0.5,0.9719614609708652\\
37,1.0,0.48598059188071974\\
37,2.0,0.24299029594035987\\
37,4.0,0.12149514797017993\\
37,nan,nan\\
38,0.0001,0.9999976024199609\\
38,0.001,0.9999976024199599\\
38,0.01,0.9999976024199602\\
38,0.1,0.9999976024199604\\
38,0.25,0.9999976024199613\\
38,0.5,0.9999976024969808\\
38,1.0,0.5897581536737098\\
38,2.0,0.2948790768368549\\
38,4.0,0.14743953841842744\\
38,nan,nan\\
39,0.0001,0.9999981988004819\\
39,0.001,0.9999981988004817\\
39,0.01,0.9999981988004821\\
39,0.1,0.999998198800482\\
39,0.25,0.9999981988004825\\
39,0.5,0.9999981726176812\\
39,1.0,0.5929983089188996\\
39,2.0,0.2964991544594498\\
39,4.0,0.1482495772297249\\
39,nan,nan\\
40,0.0001,0.9999975088759603\\
40,0.001,0.9999975088759601\\
40,0.01,0.9999975088759602\\
40,0.1,0.9999975088759612\\
40,0.25,0.9999975088759603\\
40,0.5,0.9999974825052855\\
40,1.0,0.5442101780193617\\
40,2.0,0.27210508900968083\\
40,4.0,0.13605254450484042\\
40,nan,nan\\
41,0.0001,0.9999987970079565\\
41,0.001,0.9999987970079548\\
41,0.01,0.9999987970079561\\
41,0.1,0.999998797007956\\
41,0.25,0.999998797007955\\
41,0.5,0.999998744420649\\
41,1.0,0.6964692767905571\\
41,2.0,0.3482346383952786\\
41,4.0,0.1741173191976393\\
41,nan,nan\\
42,0.0001,0.9999986520064235\\
42,0.001,0.9999986520064226\\
42,0.01,0.9999986520064242\\
42,0.1,0.9999986520064225\\
42,0.25,0.9999986520064222\\
42,0.5,0.9999986181463261\\
42,1.0,0.6320447561340732\\
42,2.0,0.3160223780670366\\
42,4.0,0.1580111890335183\\
42,nan,nan\\
43,0.0001,0.9999977701794148\\
43,0.001,0.9999977701794145\\
43,0.01,0.9999977701794154\\
43,0.1,0.9999977701794153\\
43,0.25,0.999997770179415\\
43,0.5,0.9999977466879342\\
43,1.0,0.6299503902655302\\
43,2.0,0.3149751951327651\\
43,4.0,0.15748759756638256\\
43,nan,nan\\
44,0.0001,0.9999974176361358\\
44,0.001,0.9999974176361347\\
44,0.01,0.9999974176361365\\
44,0.1,0.9999974176361359\\
44,0.25,0.9999974176361359\\
44,0.5,0.9999973984223401\\
44,1.0,0.7282633376041422\\
44,2.0,0.3641316688020711\\
44,4.0,0.18206583440103555\\
44,nan,nan\\
46,0.0001,0.999997953159062\\
46,0.001,0.999997953159061\\
46,0.01,0.9999979531590626\\
46,0.1,0.9999979531590617\\
46,0.25,0.9999979531590618\\
46,0.5,0.9999979308656263\\
46,1.0,0.6048904883818413\\
46,2.0,0.30244524419092067\\
46,4.0,0.15122262209546034\\
46,nan,nan\\
47,0.0001,0.999997523465981\\
47,0.001,0.9999975234659791\\
47,0.01,0.9999975234659793\\
47,0.1,0.9999975234659797\\
47,0.25,0.9999975234659799\\
47,0.5,0.9999974962431564\\
47,1.0,0.6753343915695289\\
47,2.0,0.33766719578476445\\
47,4.0,0.16883359789238223\\
47,nan,nan\\
48,0.0001,0.9999971138067746\\
48,0.001,0.9999971138067741\\
48,0.01,0.9999971138067745\\
48,0.1,0.9999971138067743\\
48,0.25,0.9999971138067743\\
48,0.5,0.9691727354766111\\
48,1.0,0.4845861546215569\\
48,2.0,0.24229307731077845\\
48,4.0,0.12114653865538923\\
48,nan,nan\\
49,0.0001,0.9999992165965312\\
49,0.001,0.9999992165965301\\
49,0.01,0.9999992165965296\\
49,0.1,0.9999992165965308\\
49,0.25,0.9999992165965306\\
49,0.5,0.9999991788842586\\
49,1.0,0.6170620309056413\\
49,2.0,0.30853101545282063\\
49,4.0,0.15426550772641032\\
49,nan,nan\\
51,0.0001,0.9999992615931881\\
51,0.001,0.9999992615931895\\
51,0.01,0.9999992615931877\\
51,0.1,0.9999992615931876\\
51,0.25,0.9999992615931883\\
51,0.5,0.9999992297960396\\
51,1.0,0.6548865025692444\\
51,2.0,0.3274432512846222\\
51,4.0,0.1637216256423111\\
51,nan,nan\\
52,0.0001,0.999998399512739\\
52,0.001,0.9999983995127377\\
52,0.01,0.9999983995127384\\
52,0.1,0.9999983995127382\\
52,0.25,0.9999983995127375\\
52,0.5,0.9999983833970207\\
52,1.0,0.6868198270165139\\
52,2.0,0.34340991350825695\\
52,4.0,0.17170495675412847\\
52,nan,nan\\
53,0.0001,0.99999769043899\\
53,0.001,0.9999976904389908\\
53,0.01,0.9999976904389906\\
53,0.1,0.9999976904389903\\
53,0.25,0.9999976904389889\\
53,0.5,0.9999976836335442\\
53,1.0,0.8395926959976159\\
53,2.0,0.41979634799880794\\
53,4.0,0.20989817399940397\\
53,nan,nan\\
54,0.0001,0.9999982055579456\\
54,0.001,0.9999982055579447\\
54,0.01,0.9999982055579464\\
54,0.1,0.9999982055579457\\
54,0.25,0.9999982055579462\\
54,0.5,0.999998168357569\\
54,1.0,0.5393089736590562\\
54,2.0,0.2696544868295281\\
54,4.0,0.13482724341476404\\
54,nan,nan\\
55,0.0001,0.9999976962172633\\
55,0.001,0.9999976962172642\\
55,0.01,0.9999976962172639\\
55,0.1,0.9999976962172648\\
55,0.25,0.9999976962172635\\
55,0.5,0.9999977040330817\\
55,1.0,0.5517533696971039\\
55,2.0,0.27587668484855193\\
55,4.0,0.13793834242427597\\
55,nan,nan\\
56,0.0001,0.999998246690392\\
56,0.001,0.9999982466903914\\
56,0.01,0.9999982466903914\\
56,0.1,0.9999982466903928\\
56,0.25,0.9999982466903918\\
56,0.5,0.9999982166554905\\
56,1.0,0.659703950585435\\
56,2.0,0.3298519752927175\\
56,4.0,0.16492598764635874\\
56,nan,nan\\
57,0.0001,0.9999980507850248\\
57,0.001,0.9999980507850246\\
57,0.01,0.9999980507850245\\
57,0.1,0.999998050785025\\
57,0.25,0.9999980507850252\\
57,0.5,0.9999980480934485\\
57,1.0,0.6250066871361136\\
57,2.0,0.3125033435680568\\
57,4.0,0.1562516717840284\\
57,nan,nan\\
58,0.0001,0.9999990410595421\\
58,0.001,0.999999041059542\\
58,0.01,0.9999990410595418\\
58,0.1,0.9999990410595417\\
58,0.25,0.9999990410595415\\
58,0.5,0.9999990340882766\\
58,1.0,0.9999990391706529\\
58,2.0,0.5699113863055932\\
58,4.0,0.2849556931527966\\
58,nan,nan\\
61,0.0001,0.9999976190106739\\
61,0.001,0.9999976190106749\\
61,0.01,0.9999976190106751\\
61,0.1,0.9999976190106747\\
61,0.25,0.9999976190106742\\
61,0.5,0.9999976013155543\\
61,1.0,0.728440116952241\\
61,2.0,0.3642200584761205\\
61,4.0,0.18211002923806024\\
61,nan,nan\\
62,0.0001,0.9999983466472582\\
62,0.001,0.999998346647258\\
62,0.01,0.999998346647258\\
62,0.1,0.9999983466472584\\
62,0.25,0.9999983466472604\\
62,0.5,0.99999832799997\\
62,1.0,0.5535296527791338\\
62,2.0,0.2767648263895669\\
62,4.0,0.13838241319478345\\
62,nan,nan\\
63,0.0001,0.9999985716650143\\
63,0.001,0.9999985716650139\\
63,0.01,0.9999985716650146\\
63,0.1,0.9999985716650142\\
63,0.25,0.9999985716650154\\
63,0.5,0.9999985390163718\\
63,1.0,0.605585470536903\\
63,2.0,0.3027927352684515\\
63,4.0,0.15139636763422576\\
63,nan,nan\\
64,0.0001,0.9999991206271515\\
64,0.001,0.9999991206271529\\
64,0.01,0.9999991206271523\\
64,0.1,0.9999991206271511\\
64,0.25,0.9999991206271517\\
64,0.5,0.9999990892014482\\
64,1.0,0.7220748157858836\\
64,2.0,0.3610374078929418\\
64,4.0,0.1805187039464709\\
64,nan,nan\\
66,0.0001,0.9999981616355136\\
66,0.001,0.9999981616355134\\
66,0.01,0.9999981616355146\\
66,0.1,0.9999981616355137\\
66,0.25,0.9999981616355152\\
66,0.5,0.999998159977422\\
66,1.0,0.8994082280397158\\
66,2.0,0.4497041140198579\\
66,4.0,0.22485205700992894\\
66,nan,nan\\
67,0.0001,0.9999978427345046\\
67,0.001,0.9999978427345044\\
67,0.01,0.9999978427345056\\
67,0.1,0.9999978427345031\\
67,0.25,0.9999978427345048\\
67,0.5,0.999997837079018\\
67,1.0,0.7139200478337272\\
67,2.0,0.3569600239168636\\
67,4.0,0.1784800119584318\\
67,nan,nan\\
68,0.0001,0.9999959066483204\\
68,0.001,0.9999959066483205\\
68,0.01,0.999995906648321\\
68,0.1,0.9999959066483205\\
68,0.25,0.9999959066483207\\
68,0.5,0.9999959155870648\\
68,1.0,0.5508054192117025\\
68,2.0,0.2754027096058512\\
68,4.0,0.1377013548029256\\
68,nan,nan\\
69,0.0001,0.9999980812669381\\
69,0.001,0.9999980812669387\\
69,0.01,0.9999980812669377\\
69,0.1,0.9999980812669387\\
69,0.25,0.9999980812669381\\
69,0.5,0.9999980650598806\\
69,1.0,0.6563935912157196\\
69,2.0,0.3281967956078598\\
69,4.0,0.1640983978039299\\
69,nan,nan\\
70,0.0001,0.9999989683462962\\
70,0.001,0.9999989683462951\\
70,0.01,0.9999989683462946\\
70,0.1,0.9999989683462941\\
70,0.25,0.9999989683462952\\
70,0.5,0.9999989523952351\\
70,1.0,0.8756533481106461\\
70,2.0,0.43782667405532305\\
70,4.0,0.21891333702766153\\
70,nan,nan\\
71,0.0001,0.9999983087396516\\
71,0.001,0.9999983087396517\\
71,0.01,0.9999983087396503\\
71,0.1,0.9999983087396508\\
71,0.25,0.9999983087396519\\
71,0.5,0.9999983055407624\\
71,1.0,0.7096626405976876\\
71,2.0,0.3548313202988438\\
71,4.0,0.1774156601494219\\
71,nan,nan\\
73,0.0001,0.9999997238073337\\
73,0.001,0.9999997238073347\\
73,0.01,0.999999723807335\\
73,0.1,0.9999997238073347\\
73,0.25,0.9999997238073338\\
73,0.5,0.9999996775288832\\
73,1.0,0.5783639087515947\\
73,2.0,0.28918195437579736\\
73,4.0,0.14459097718789868\\
73,nan,nan\\
74,0.0001,0.9999971931192384\\
74,0.001,0.9999971931192387\\
74,0.01,0.9999971931192395\\
74,0.1,0.9999971931192388\\
74,0.25,0.99999719311924\\
74,0.5,0.9999971780647255\\
74,1.0,0.5716634304575777\\
74,2.0,0.28583171522878886\\
74,4.0,0.14291585761439443\\
74,nan,nan\\
77,0.0001,0.9999973153958035\\
77,0.001,0.9999973153958037\\
77,0.01,0.9999973153958021\\
77,0.1,0.9999973153958028\\
77,0.25,0.999997315395803\\
77,0.5,0.9999973140817625\\
77,1.0,0.6333217463446627\\
77,2.0,0.3166608731723313\\
77,4.0,0.15833043658616566\\
77,nan,nan\\
78,0.0001,0.9999992651834013\\
78,0.001,0.9999992651834022\\
78,0.01,0.999999265183402\\
78,0.1,0.999999265183402\\
78,0.25,0.9999992651834019\\
78,0.5,0.9999992420373767\\
78,1.0,0.9999991449212268\\
78,2.0,0.5690836040038267\\
78,4.0,0.28454180200191337\\
78,nan,nan\\
79,0.0001,0.9999983926898686\\
79,0.001,0.9999983926898676\\
79,0.01,0.9999983926898685\\
79,0.1,0.9999983926898668\\
79,0.25,0.9999983926898678\\
79,0.5,0.9999983652242496\\
79,1.0,0.6418495502412744\\
79,2.0,0.3209247751206372\\
79,4.0,0.1604623875603186\\
79,nan,nan\\
80,0.0001,0.9999997129845936\\
80,0.001,0.999999712984594\\
80,0.01,0.9999997129845936\\
80,0.1,0.9999997129845917\\
80,0.25,0.9999997129845943\\
80,0.5,0.9999996770044991\\
80,1.0,0.6927584005763544\\
80,2.0,0.3463792002881772\\
80,4.0,0.1731896001440886\\
80,nan,nan\\
81,0.0001,0.9999989250212342\\
81,0.001,0.9999989250212347\\
81,0.01,0.9999989250212329\\
81,0.1,0.999998925021233\\
81,0.25,0.9999989250212328\\
81,0.5,0.9999988957304607\\
81,1.0,0.8201040815464715\\
81,2.0,0.41005204077323576\\
81,4.0,0.20502602038661788\\
81,nan,nan\\
82,0.0001,0.9999985209316009\\
82,0.001,0.9999985209316019\\
82,0.01,0.9999985209316015\\
82,0.1,0.9999985209316026\\
82,0.25,0.9999985209316015\\
82,0.5,0.9999984946907148\\
82,1.0,0.850569331684953\\
82,2.0,0.4252846658424765\\
82,4.0,0.21264233292123824\\
82,nan,nan\\
83,0.0001,0.999999122956182\\
83,0.001,0.9999991229561813\\
83,0.01,0.9999991229561815\\
83,0.1,0.9999991229561819\\
83,0.25,0.9999991229561811\\
83,0.5,0.9999990956859527\\
83,1.0,0.9029713830192627\\
83,2.0,0.45148569150963136\\
83,4.0,0.22574284575481568\\
83,nan,nan\\
84,0.0001,0.9999986815834848\\
84,0.001,0.9999986815834838\\
84,0.01,0.9999986815834827\\
84,0.1,0.9999986815834853\\
84,0.25,0.999998681583485\\
84,0.5,0.999998654049262\\
84,1.0,0.7051209929796667\\
84,2.0,0.35256049648983334\\
84,4.0,0.17628024824491667\\
84,nan,nan\\
85,0.0001,0.9999983625306433\\
85,0.001,0.9999983625306443\\
85,0.01,0.9999983625306451\\
85,0.1,0.9999983625306437\\
85,0.25,0.9999983625306446\\
85,0.5,0.9999983362388946\\
85,1.0,0.5618378078393563\\
85,2.0,0.28091890391967816\\
85,4.0,0.14045945195983908\\
85,nan,nan\\
86,0.0001,0.9999971373098377\\
86,0.001,0.9999971373098376\\
86,0.01,0.9999971373098371\\
86,0.1,0.9999971373098383\\
86,0.25,0.999997137309838\\
86,0.5,0.9999971145267944\\
86,1.0,0.5396588595356911\\
86,2.0,0.2698294297678456\\
86,4.0,0.1349147148839228\\
86,nan,nan\\
87,0.0001,0.9999963875054444\\
87,0.001,0.999996387505444\\
87,0.01,0.9999963875054444\\
87,0.1,0.9999963875054448\\
87,0.25,0.9999963875054436\\
87,0.5,0.9999963919972329\\
87,1.0,0.6276527495253307\\
87,2.0,0.31382637476266534\\
87,4.0,0.15691318738133267\\
87,nan,nan\\
};
\addplot[native, line width=0.65pt, unbounded coords=jump] table[x=C, y=kappa, col sep=comma, row sep=crcr] {
candidate,C,kappa\\
26,0.0001,1.041394421405705\\
26,0.001,1.0413944214057038\\
26,0.01,1.0413944214057036\\
26,0.1,1.0413944214057047\\
26,0.25,1.0413944214057043\\
26,0.5,1.039929417011008\\
26,1.0,0.6420252344558547\\
26,2.0,0.32101261722792734\\
26,4.0,0.16050630861396367\\
26,nan,nan\\
35,0.0001,1.0324653186293193\\
35,0.001,1.0324653186293202\\
35,0.01,1.032465318629317\\
35,0.1,1.0324653186293187\\
35,0.25,1.0324653186293196\\
35,0.5,1.0316082805866862\\
35,1.0,0.7218544674425208\\
35,2.0,0.3609272337212604\\
35,4.0,0.1804636168606302\\
35,nan,nan\\
45,0.0001,1.046143359721761\\
45,0.001,1.0461433597217602\\
45,0.01,1.0461433597217615\\
45,0.1,1.046143359721759\\
45,0.25,1.0461433597217609\\
45,0.5,1.0451711030649067\\
45,1.0,0.5658988934147263\\
45,2.0,0.28294944670736316\\
45,4.0,0.14147472335368158\\
45,nan,nan\\
50,0.0001,1.0477562553710593\\
50,0.001,1.0477562553710589\\
50,0.01,1.0477562553710609\\
50,0.1,1.0477562553710589\\
50,0.25,1.0477562553710587\\
50,0.5,1.0460359276211135\\
50,1.0,0.6084970983946533\\
50,2.0,0.30424854919732663\\
50,4.0,0.15212427459866332\\
50,nan,nan\\
59,0.0001,1.0602568221668587\\
59,0.001,1.0602568221668582\\
59,0.01,1.0602568221668576\\
59,0.1,1.0602568221668593\\
59,0.25,1.060256822166858\\
59,0.5,1.0579779299175245\\
59,1.0,0.7837038520858092\\
59,2.0,0.3918519260429046\\
59,4.0,0.1959259630214523\\
59,nan,nan\\
60,0.0001,1.046193794533972\\
60,0.001,1.0461937945339732\\
60,0.01,1.0461937945339723\\
60,0.1,1.0461937945339728\\
60,0.25,1.0461937945339728\\
60,0.5,1.0448331782605893\\
60,1.0,0.7347462217269578\\
60,2.0,0.3673731108634789\\
60,4.0,0.18368655543173945\\
60,nan,nan\\
65,0.0001,1.0614930615798064\\
65,0.001,1.0614930615798082\\
65,0.01,1.0614930615798064\\
65,0.1,1.0614930615798084\\
65,0.25,1.061493061579806\\
65,0.5,1.060286444019637\\
65,1.0,0.7394787872910571\\
65,2.0,0.36973939364552855\\
65,4.0,0.18486969682276427\\
65,nan,nan\\
72,0.0001,1.0438999037859669\\
72,0.001,1.043899903785965\\
72,0.01,1.043899903785966\\
72,0.1,1.0438999037859666\\
72,0.25,1.0438999037859649\\
72,0.5,1.0428162544559698\\
72,1.0,0.7843292170465982\\
72,2.0,0.3921646085232991\\
72,4.0,0.19608230426164955\\
72,nan,nan\\
75,0.0001,1.06525795957682\\
75,0.001,1.0652579595768203\\
75,0.01,1.0652579595768195\\
75,0.1,1.0652579595768197\\
75,0.25,1.0652579595768203\\
75,0.5,1.0640552302235247\\
75,1.0,0.7131179984447849\\
75,2.0,0.3565589992223924\\
75,4.0,0.1782794996111962\\
75,nan,nan\\
76,0.0001,1.0144871510188735\\
76,0.001,1.014487151018875\\
76,0.01,1.0144871510188755\\
76,0.1,1.0144871510188753\\
76,0.25,1.014487151018876\\
76,0.5,1.0129599756492444\\
76,1.0,0.9282782024888052\\
76,2.0,0.4641391012444026\\
76,4.0,0.2320695506222013\\
76,nan,nan\\
};
\addplot[only marks, mark=*, mark size=2.2pt, native, mark options={draw=white, line width=0.6pt}] coordinates {(0.001,1.065258) (0.25,1.065258)};
\node[note, anchor=south west] at (axis cs:1.05e-4,1.082) {\textbf{10 of 64} exceed $C$ by 1.4--6.5\%};
\node[note, anchor=south east] at (axis cs:5.3,1.082) {dots: top candidate revalidated natively, $\kappa{=}1.0653$};
\node[note, anchor=south east] at (axis cs:5.3,1.001) {bound $C$};
\end{axis}
\begin{axis}[paper, name=bot, clip=true, clip mode=individual,
  at={($(top.south west)+(0,-0.16cm)$)}, anchor=north west,
  width=9.4cm, height=1.7cm, scale only axis,
  xmode=log, xmin=7e-5, xmax=5.5,
  xtick={1e-4,1e-3,1e-2,1e-1,1}, xticklabels={$10^{-4}$,$10^{-3}$,$10^{-2}$,$10^{-1}$,$1$},
  extra x ticks={0.25,4}, extra x tick labels={$0.25$,$4$},
  extra x tick style={grid=none},
  ymin=0, ymax=0.994, ytick={0,0.25,0.5,0.75},
  yticklabel style={/pgf/number format/fixed, /pgf/number format/precision=2},
  ymajorgrids,
  xlabel={Clipping radius $C$ (log scale)},
]
\fill[wash] (axis cs:7e-5,0) rectangle (axis cs:0.25,0.994);
\addplot[faint, line width=0.45pt, unbounded coords=jump] table[x=C, y=kappa, col sep=comma, row sep=crcr] {
candidate,C,kappa\\
24,0.0001,0.9999973708636837\\
24,0.001,0.9999973708636851\\
24,0.01,0.9999973708636846\\
24,0.1,0.9999973708636841\\
24,0.25,0.9999973708636861\\
24,0.5,0.9999973445230439\\
24,1.0,0.5347295375545472\\
24,2.0,0.2673647687772736\\
24,4.0,0.1336823843886368\\
24,nan,nan\\
25,0.0001,0.9999985062968676\\
25,0.001,0.9999985062968694\\
25,0.01,0.9999985062968673\\
25,0.1,0.9999985062968682\\
25,0.25,0.9999985062968669\\
25,0.5,0.9999984681413554\\
25,1.0,0.5672308555343831\\
25,2.0,0.28361542776719156\\
25,4.0,0.14180771388359578\\
25,nan,nan\\
27,0.0001,0.999996724697214\\
27,0.001,0.9999967246972137\\
27,0.01,0.9999967246972137\\
27,0.1,0.9999967246972143\\
27,0.25,0.9999967246972133\\
27,0.5,0.9754419512151915\\
27,1.0,0.48772070114125404\\
27,2.0,0.24386035057062702\\
27,4.0,0.12193017528531351\\
27,nan,nan\\
28,0.0001,0.9999981632548344\\
28,0.001,0.9999981632548344\\
28,0.01,0.9999981632548347\\
28,0.1,0.9999981632548339\\
28,0.25,0.9999981632548343\\
28,0.5,0.999998120777037\\
28,1.0,0.6092633930875981\\
28,2.0,0.30463169654379907\\
28,4.0,0.15231584827189953\\
28,nan,nan\\
29,0.0001,0.9999976969124461\\
29,0.001,0.9999976969124469\\
29,0.01,0.9999976969124467\\
29,0.1,0.9999976969124463\\
29,0.25,0.9999976969124469\\
29,0.5,0.9999976520437359\\
29,1.0,0.6055618202276727\\
29,2.0,0.30278091011383634\\
29,4.0,0.15139045505691817\\
29,nan,nan\\
30,0.0001,0.9999975290595678\\
30,0.001,0.9999975290595662\\
30,0.01,0.9999975290595672\\
30,0.1,0.9999975290595676\\
30,0.25,0.9999975290595675\\
30,0.5,0.9999974981435457\\
30,1.0,0.6246483026938584\\
30,2.0,0.3123241513469292\\
30,4.0,0.1561620756734646\\
30,nan,nan\\
31,0.0001,0.9999981722033685\\
31,0.001,0.9999981722033685\\
31,0.01,0.9999981722033672\\
31,0.1,0.999998172203368\\
31,0.25,0.9999981722033684\\
31,0.5,0.9999981226395972\\
31,1.0,0.526259355638783\\
31,2.0,0.2631296778193915\\
31,4.0,0.13156483890969575\\
31,nan,nan\\
32,0.0001,0.999996957741305\\
32,0.001,0.999996957741305\\
32,0.01,0.9999969577413069\\
32,0.1,0.999996957741306\\
32,0.25,0.9999969577413058\\
32,0.5,0.9999969276567893\\
32,1.0,0.640132969444828\\
32,2.0,0.320066484722414\\
32,4.0,0.160033242361207\\
32,nan,nan\\
33,0.0001,1.0000000461266045\\
33,0.001,1.0000000461266056\\
33,0.01,1.0000000461266048\\
33,0.1,1.0000000461266043\\
33,0.25,1.0000000461266036\\
33,0.5,0.9999999821715908\\
33,1.0,0.5599606374668892\\
33,2.0,0.2799803187334446\\
33,4.0,0.1399901593667223\\
33,nan,nan\\
34,0.0001,0.9999984200634098\\
34,0.001,0.9999984200634093\\
34,0.01,0.9999984200634099\\
34,0.1,0.999998420063411\\
34,0.25,0.9999984200634093\\
34,0.5,0.9999983790130175\\
34,1.0,0.5599851075788533\\
34,2.0,0.27999255378942667\\
34,4.0,0.13999627689471333\\
34,nan,nan\\
36,0.0001,0.999998479259472\\
36,0.001,0.9999984792594708\\
36,0.01,0.9999984792594717\\
36,0.1,0.9999984792594718\\
36,0.25,0.9999984792594707\\
36,0.5,0.9999984285593985\\
36,1.0,0.581410361047393\\
36,2.0,0.2907051805236965\\
36,4.0,0.14535259026184824\\
36,nan,nan\\
37,0.0001,0.9999979102196204\\
37,0.001,0.9999979102196198\\
37,0.01,0.9999979102196191\\
37,0.1,0.999997910219619\\
37,0.25,0.9999979102196198\\
37,0.5,0.9719614609708652\\
37,1.0,0.48598059188071974\\
37,2.0,0.24299029594035987\\
37,4.0,0.12149514797017993\\
37,nan,nan\\
38,0.0001,0.9999976024199609\\
38,0.001,0.9999976024199599\\
38,0.01,0.9999976024199602\\
38,0.1,0.9999976024199604\\
38,0.25,0.9999976024199613\\
38,0.5,0.9999976024969808\\
38,1.0,0.5897581536737098\\
38,2.0,0.2948790768368549\\
38,4.0,0.14743953841842744\\
38,nan,nan\\
39,0.0001,0.9999981988004819\\
39,0.001,0.9999981988004817\\
39,0.01,0.9999981988004821\\
39,0.1,0.999998198800482\\
39,0.25,0.9999981988004825\\
39,0.5,0.9999981726176812\\
39,1.0,0.5929983089188996\\
39,2.0,0.2964991544594498\\
39,4.0,0.1482495772297249\\
39,nan,nan\\
40,0.0001,0.9999975088759603\\
40,0.001,0.9999975088759601\\
40,0.01,0.9999975088759602\\
40,0.1,0.9999975088759612\\
40,0.25,0.9999975088759603\\
40,0.5,0.9999974825052855\\
40,1.0,0.5442101780193617\\
40,2.0,0.27210508900968083\\
40,4.0,0.13605254450484042\\
40,nan,nan\\
41,0.0001,0.9999987970079565\\
41,0.001,0.9999987970079548\\
41,0.01,0.9999987970079561\\
41,0.1,0.999998797007956\\
41,0.25,0.999998797007955\\
41,0.5,0.999998744420649\\
41,1.0,0.6964692767905571\\
41,2.0,0.3482346383952786\\
41,4.0,0.1741173191976393\\
41,nan,nan\\
42,0.0001,0.9999986520064235\\
42,0.001,0.9999986520064226\\
42,0.01,0.9999986520064242\\
42,0.1,0.9999986520064225\\
42,0.25,0.9999986520064222\\
42,0.5,0.9999986181463261\\
42,1.0,0.6320447561340732\\
42,2.0,0.3160223780670366\\
42,4.0,0.1580111890335183\\
42,nan,nan\\
43,0.0001,0.9999977701794148\\
43,0.001,0.9999977701794145\\
43,0.01,0.9999977701794154\\
43,0.1,0.9999977701794153\\
43,0.25,0.999997770179415\\
43,0.5,0.9999977466879342\\
43,1.0,0.6299503902655302\\
43,2.0,0.3149751951327651\\
43,4.0,0.15748759756638256\\
43,nan,nan\\
44,0.0001,0.9999974176361358\\
44,0.001,0.9999974176361347\\
44,0.01,0.9999974176361365\\
44,0.1,0.9999974176361359\\
44,0.25,0.9999974176361359\\
44,0.5,0.9999973984223401\\
44,1.0,0.7282633376041422\\
44,2.0,0.3641316688020711\\
44,4.0,0.18206583440103555\\
44,nan,nan\\
46,0.0001,0.999997953159062\\
46,0.001,0.999997953159061\\
46,0.01,0.9999979531590626\\
46,0.1,0.9999979531590617\\
46,0.25,0.9999979531590618\\
46,0.5,0.9999979308656263\\
46,1.0,0.6048904883818413\\
46,2.0,0.30244524419092067\\
46,4.0,0.15122262209546034\\
46,nan,nan\\
47,0.0001,0.999997523465981\\
47,0.001,0.9999975234659791\\
47,0.01,0.9999975234659793\\
47,0.1,0.9999975234659797\\
47,0.25,0.9999975234659799\\
47,0.5,0.9999974962431564\\
47,1.0,0.6753343915695289\\
47,2.0,0.33766719578476445\\
47,4.0,0.16883359789238223\\
47,nan,nan\\
48,0.0001,0.9999971138067746\\
48,0.001,0.9999971138067741\\
48,0.01,0.9999971138067745\\
48,0.1,0.9999971138067743\\
48,0.25,0.9999971138067743\\
48,0.5,0.9691727354766111\\
48,1.0,0.4845861546215569\\
48,2.0,0.24229307731077845\\
48,4.0,0.12114653865538923\\
48,nan,nan\\
49,0.0001,0.9999992165965312\\
49,0.001,0.9999992165965301\\
49,0.01,0.9999992165965296\\
49,0.1,0.9999992165965308\\
49,0.25,0.9999992165965306\\
49,0.5,0.9999991788842586\\
49,1.0,0.6170620309056413\\
49,2.0,0.30853101545282063\\
49,4.0,0.15426550772641032\\
49,nan,nan\\
51,0.0001,0.9999992615931881\\
51,0.001,0.9999992615931895\\
51,0.01,0.9999992615931877\\
51,0.1,0.9999992615931876\\
51,0.25,0.9999992615931883\\
51,0.5,0.9999992297960396\\
51,1.0,0.6548865025692444\\
51,2.0,0.3274432512846222\\
51,4.0,0.1637216256423111\\
51,nan,nan\\
52,0.0001,0.999998399512739\\
52,0.001,0.9999983995127377\\
52,0.01,0.9999983995127384\\
52,0.1,0.9999983995127382\\
52,0.25,0.9999983995127375\\
52,0.5,0.9999983833970207\\
52,1.0,0.6868198270165139\\
52,2.0,0.34340991350825695\\
52,4.0,0.17170495675412847\\
52,nan,nan\\
53,0.0001,0.99999769043899\\
53,0.001,0.9999976904389908\\
53,0.01,0.9999976904389906\\
53,0.1,0.9999976904389903\\
53,0.25,0.9999976904389889\\
53,0.5,0.9999976836335442\\
53,1.0,0.8395926959976159\\
53,2.0,0.41979634799880794\\
53,4.0,0.20989817399940397\\
53,nan,nan\\
54,0.0001,0.9999982055579456\\
54,0.001,0.9999982055579447\\
54,0.01,0.9999982055579464\\
54,0.1,0.9999982055579457\\
54,0.25,0.9999982055579462\\
54,0.5,0.999998168357569\\
54,1.0,0.5393089736590562\\
54,2.0,0.2696544868295281\\
54,4.0,0.13482724341476404\\
54,nan,nan\\
55,0.0001,0.9999976962172633\\
55,0.001,0.9999976962172642\\
55,0.01,0.9999976962172639\\
55,0.1,0.9999976962172648\\
55,0.25,0.9999976962172635\\
55,0.5,0.9999977040330817\\
55,1.0,0.5517533696971039\\
55,2.0,0.27587668484855193\\
55,4.0,0.13793834242427597\\
55,nan,nan\\
56,0.0001,0.999998246690392\\
56,0.001,0.9999982466903914\\
56,0.01,0.9999982466903914\\
56,0.1,0.9999982466903928\\
56,0.25,0.9999982466903918\\
56,0.5,0.9999982166554905\\
56,1.0,0.659703950585435\\
56,2.0,0.3298519752927175\\
56,4.0,0.16492598764635874\\
56,nan,nan\\
57,0.0001,0.9999980507850248\\
57,0.001,0.9999980507850246\\
57,0.01,0.9999980507850245\\
57,0.1,0.999998050785025\\
57,0.25,0.9999980507850252\\
57,0.5,0.9999980480934485\\
57,1.0,0.6250066871361136\\
57,2.0,0.3125033435680568\\
57,4.0,0.1562516717840284\\
57,nan,nan\\
58,0.0001,0.9999990410595421\\
58,0.001,0.999999041059542\\
58,0.01,0.9999990410595418\\
58,0.1,0.9999990410595417\\
58,0.25,0.9999990410595415\\
58,0.5,0.9999990340882766\\
58,1.0,0.9999990391706529\\
58,2.0,0.5699113863055932\\
58,4.0,0.2849556931527966\\
58,nan,nan\\
61,0.0001,0.9999976190106739\\
61,0.001,0.9999976190106749\\
61,0.01,0.9999976190106751\\
61,0.1,0.9999976190106747\\
61,0.25,0.9999976190106742\\
61,0.5,0.9999976013155543\\
61,1.0,0.728440116952241\\
61,2.0,0.3642200584761205\\
61,4.0,0.18211002923806024\\
61,nan,nan\\
62,0.0001,0.9999983466472582\\
62,0.001,0.999998346647258\\
62,0.01,0.999998346647258\\
62,0.1,0.9999983466472584\\
62,0.25,0.9999983466472604\\
62,0.5,0.99999832799997\\
62,1.0,0.5535296527791338\\
62,2.0,0.2767648263895669\\
62,4.0,0.13838241319478345\\
62,nan,nan\\
63,0.0001,0.9999985716650143\\
63,0.001,0.9999985716650139\\
63,0.01,0.9999985716650146\\
63,0.1,0.9999985716650142\\
63,0.25,0.9999985716650154\\
63,0.5,0.9999985390163718\\
63,1.0,0.605585470536903\\
63,2.0,0.3027927352684515\\
63,4.0,0.15139636763422576\\
63,nan,nan\\
64,0.0001,0.9999991206271515\\
64,0.001,0.9999991206271529\\
64,0.01,0.9999991206271523\\
64,0.1,0.9999991206271511\\
64,0.25,0.9999991206271517\\
64,0.5,0.9999990892014482\\
64,1.0,0.7220748157858836\\
64,2.0,0.3610374078929418\\
64,4.0,0.1805187039464709\\
64,nan,nan\\
66,0.0001,0.9999981616355136\\
66,0.001,0.9999981616355134\\
66,0.01,0.9999981616355146\\
66,0.1,0.9999981616355137\\
66,0.25,0.9999981616355152\\
66,0.5,0.999998159977422\\
66,1.0,0.8994082280397158\\
66,2.0,0.4497041140198579\\
66,4.0,0.22485205700992894\\
66,nan,nan\\
67,0.0001,0.9999978427345046\\
67,0.001,0.9999978427345044\\
67,0.01,0.9999978427345056\\
67,0.1,0.9999978427345031\\
67,0.25,0.9999978427345048\\
67,0.5,0.999997837079018\\
67,1.0,0.7139200478337272\\
67,2.0,0.3569600239168636\\
67,4.0,0.1784800119584318\\
67,nan,nan\\
68,0.0001,0.9999959066483204\\
68,0.001,0.9999959066483205\\
68,0.01,0.999995906648321\\
68,0.1,0.9999959066483205\\
68,0.25,0.9999959066483207\\
68,0.5,0.9999959155870648\\
68,1.0,0.5508054192117025\\
68,2.0,0.2754027096058512\\
68,4.0,0.1377013548029256\\
68,nan,nan\\
69,0.0001,0.9999980812669381\\
69,0.001,0.9999980812669387\\
69,0.01,0.9999980812669377\\
69,0.1,0.9999980812669387\\
69,0.25,0.9999980812669381\\
69,0.5,0.9999980650598806\\
69,1.0,0.6563935912157196\\
69,2.0,0.3281967956078598\\
69,4.0,0.1640983978039299\\
69,nan,nan\\
70,0.0001,0.9999989683462962\\
70,0.001,0.9999989683462951\\
70,0.01,0.9999989683462946\\
70,0.1,0.9999989683462941\\
70,0.25,0.9999989683462952\\
70,0.5,0.9999989523952351\\
70,1.0,0.8756533481106461\\
70,2.0,0.43782667405532305\\
70,4.0,0.21891333702766153\\
70,nan,nan\\
71,0.0001,0.9999983087396516\\
71,0.001,0.9999983087396517\\
71,0.01,0.9999983087396503\\
71,0.1,0.9999983087396508\\
71,0.25,0.9999983087396519\\
71,0.5,0.9999983055407624\\
71,1.0,0.7096626405976876\\
71,2.0,0.3548313202988438\\
71,4.0,0.1774156601494219\\
71,nan,nan\\
73,0.0001,0.9999997238073337\\
73,0.001,0.9999997238073347\\
73,0.01,0.999999723807335\\
73,0.1,0.9999997238073347\\
73,0.25,0.9999997238073338\\
73,0.5,0.9999996775288832\\
73,1.0,0.5783639087515947\\
73,2.0,0.28918195437579736\\
73,4.0,0.14459097718789868\\
73,nan,nan\\
74,0.0001,0.9999971931192384\\
74,0.001,0.9999971931192387\\
74,0.01,0.9999971931192395\\
74,0.1,0.9999971931192388\\
74,0.25,0.99999719311924\\
74,0.5,0.9999971780647255\\
74,1.0,0.5716634304575777\\
74,2.0,0.28583171522878886\\
74,4.0,0.14291585761439443\\
74,nan,nan\\
77,0.0001,0.9999973153958035\\
77,0.001,0.9999973153958037\\
77,0.01,0.9999973153958021\\
77,0.1,0.9999973153958028\\
77,0.25,0.999997315395803\\
77,0.5,0.9999973140817625\\
77,1.0,0.6333217463446627\\
77,2.0,0.3166608731723313\\
77,4.0,0.15833043658616566\\
77,nan,nan\\
78,0.0001,0.9999992651834013\\
78,0.001,0.9999992651834022\\
78,0.01,0.999999265183402\\
78,0.1,0.999999265183402\\
78,0.25,0.9999992651834019\\
78,0.5,0.9999992420373767\\
78,1.0,0.9999991449212268\\
78,2.0,0.5690836040038267\\
78,4.0,0.28454180200191337\\
78,nan,nan\\
79,0.0001,0.9999983926898686\\
79,0.001,0.9999983926898676\\
79,0.01,0.9999983926898685\\
79,0.1,0.9999983926898668\\
79,0.25,0.9999983926898678\\
79,0.5,0.9999983652242496\\
79,1.0,0.6418495502412744\\
79,2.0,0.3209247751206372\\
79,4.0,0.1604623875603186\\
79,nan,nan\\
80,0.0001,0.9999997129845936\\
80,0.001,0.999999712984594\\
80,0.01,0.9999997129845936\\
80,0.1,0.9999997129845917\\
80,0.25,0.9999997129845943\\
80,0.5,0.9999996770044991\\
80,1.0,0.6927584005763544\\
80,2.0,0.3463792002881772\\
80,4.0,0.1731896001440886\\
80,nan,nan\\
81,0.0001,0.9999989250212342\\
81,0.001,0.9999989250212347\\
81,0.01,0.9999989250212329\\
81,0.1,0.999998925021233\\
81,0.25,0.9999989250212328\\
81,0.5,0.9999988957304607\\
81,1.0,0.8201040815464715\\
81,2.0,0.41005204077323576\\
81,4.0,0.20502602038661788\\
81,nan,nan\\
82,0.0001,0.9999985209316009\\
82,0.001,0.9999985209316019\\
82,0.01,0.9999985209316015\\
82,0.1,0.9999985209316026\\
82,0.25,0.9999985209316015\\
82,0.5,0.9999984946907148\\
82,1.0,0.850569331684953\\
82,2.0,0.4252846658424765\\
82,4.0,0.21264233292123824\\
82,nan,nan\\
83,0.0001,0.999999122956182\\
83,0.001,0.9999991229561813\\
83,0.01,0.9999991229561815\\
83,0.1,0.9999991229561819\\
83,0.25,0.9999991229561811\\
83,0.5,0.9999990956859527\\
83,1.0,0.9029713830192627\\
83,2.0,0.45148569150963136\\
83,4.0,0.22574284575481568\\
83,nan,nan\\
84,0.0001,0.9999986815834848\\
84,0.001,0.9999986815834838\\
84,0.01,0.9999986815834827\\
84,0.1,0.9999986815834853\\
84,0.25,0.999998681583485\\
84,0.5,0.999998654049262\\
84,1.0,0.7051209929796667\\
84,2.0,0.35256049648983334\\
84,4.0,0.17628024824491667\\
84,nan,nan\\
85,0.0001,0.9999983625306433\\
85,0.001,0.9999983625306443\\
85,0.01,0.9999983625306451\\
85,0.1,0.9999983625306437\\
85,0.25,0.9999983625306446\\
85,0.5,0.9999983362388946\\
85,1.0,0.5618378078393563\\
85,2.0,0.28091890391967816\\
85,4.0,0.14045945195983908\\
85,nan,nan\\
86,0.0001,0.9999971373098377\\
86,0.001,0.9999971373098376\\
86,0.01,0.9999971373098371\\
86,0.1,0.9999971373098383\\
86,0.25,0.999997137309838\\
86,0.5,0.9999971145267944\\
86,1.0,0.5396588595356911\\
86,2.0,0.2698294297678456\\
86,4.0,0.1349147148839228\\
86,nan,nan\\
87,0.0001,0.9999963875054444\\
87,0.001,0.999996387505444\\
87,0.01,0.9999963875054444\\
87,0.1,0.9999963875054448\\
87,0.25,0.9999963875054436\\
87,0.5,0.9999963919972329\\
87,1.0,0.6276527495253307\\
87,2.0,0.31382637476266534\\
87,4.0,0.15691318738133267\\
87,nan,nan\\
};
\addplot[native, line width=0.65pt, unbounded coords=jump] table[x=C, y=kappa, col sep=comma, row sep=crcr] {
candidate,C,kappa\\
26,0.0001,1.041394421405705\\
26,0.001,1.0413944214057038\\
26,0.01,1.0413944214057036\\
26,0.1,1.0413944214057047\\
26,0.25,1.0413944214057043\\
26,0.5,1.039929417011008\\
26,1.0,0.6420252344558547\\
26,2.0,0.32101261722792734\\
26,4.0,0.16050630861396367\\
26,nan,nan\\
35,0.0001,1.0324653186293193\\
35,0.001,1.0324653186293202\\
35,0.01,1.032465318629317\\
35,0.1,1.0324653186293187\\
35,0.25,1.0324653186293196\\
35,0.5,1.0316082805866862\\
35,1.0,0.7218544674425208\\
35,2.0,0.3609272337212604\\
35,4.0,0.1804636168606302\\
35,nan,nan\\
45,0.0001,1.046143359721761\\
45,0.001,1.0461433597217602\\
45,0.01,1.0461433597217615\\
45,0.1,1.046143359721759\\
45,0.25,1.0461433597217609\\
45,0.5,1.0451711030649067\\
45,1.0,0.5658988934147263\\
45,2.0,0.28294944670736316\\
45,4.0,0.14147472335368158\\
45,nan,nan\\
50,0.0001,1.0477562553710593\\
50,0.001,1.0477562553710589\\
50,0.01,1.0477562553710609\\
50,0.1,1.0477562553710589\\
50,0.25,1.0477562553710587\\
50,0.5,1.0460359276211135\\
50,1.0,0.6084970983946533\\
50,2.0,0.30424854919732663\\
50,4.0,0.15212427459866332\\
50,nan,nan\\
59,0.0001,1.0602568221668587\\
59,0.001,1.0602568221668582\\
59,0.01,1.0602568221668576\\
59,0.1,1.0602568221668593\\
59,0.25,1.060256822166858\\
59,0.5,1.0579779299175245\\
59,1.0,0.7837038520858092\\
59,2.0,0.3918519260429046\\
59,4.0,0.1959259630214523\\
59,nan,nan\\
60,0.0001,1.046193794533972\\
60,0.001,1.0461937945339732\\
60,0.01,1.0461937945339723\\
60,0.1,1.0461937945339728\\
60,0.25,1.0461937945339728\\
60,0.5,1.0448331782605893\\
60,1.0,0.7347462217269578\\
60,2.0,0.3673731108634789\\
60,4.0,0.18368655543173945\\
60,nan,nan\\
65,0.0001,1.0614930615798064\\
65,0.001,1.0614930615798082\\
65,0.01,1.0614930615798064\\
65,0.1,1.0614930615798084\\
65,0.25,1.061493061579806\\
65,0.5,1.060286444019637\\
65,1.0,0.7394787872910571\\
65,2.0,0.36973939364552855\\
65,4.0,0.18486969682276427\\
65,nan,nan\\
72,0.0001,1.0438999037859669\\
72,0.001,1.043899903785965\\
72,0.01,1.043899903785966\\
72,0.1,1.0438999037859666\\
72,0.25,1.0438999037859649\\
72,0.5,1.0428162544559698\\
72,1.0,0.7843292170465982\\
72,2.0,0.3921646085232991\\
72,4.0,0.19608230426164955\\
72,nan,nan\\
75,0.0001,1.06525795957682\\
75,0.001,1.0652579595768203\\
75,0.01,1.0652579595768195\\
75,0.1,1.0652579595768197\\
75,0.25,1.0652579595768203\\
75,0.5,1.0640552302235247\\
75,1.0,0.7131179984447849\\
75,2.0,0.3565589992223924\\
75,4.0,0.1782794996111962\\
75,nan,nan\\
76,0.0001,1.0144871510188735\\
76,0.001,1.014487151018875\\
76,0.01,1.0144871510188755\\
76,0.1,1.0144871510188753\\
76,0.25,1.014487151018876\\
76,0.5,1.0129599756492444\\
76,1.0,0.9282782024888052\\
76,2.0,0.4641391012444026\\
76,4.0,0.2320695506222013\\
76,nan,nan\\
};
\draw[ink2, line width=0.5pt] (axis cs:1,0) -- (axis cs:1,0.994);
\node[note, anchor=north west, align=left] at (axis cs:1.05e-4,0.93) {shaded: every gradient is clipped,\\ so $\kappa$ does not depend on $C$};
\node[note, anchor=south east, align=right, inner sep=1pt] (scr) at (axis cs:0.2,0.05) {original screen at $C{=}1$:\\ none of 64 exceed $C$};
\draw[-{Latex[length=1.2mm]}, ink2, line width=0.4pt] (scr.east) -- (axis cs:0.94,0.12);
\end{axis}
\node[rotate=90, font=\scriptsize, text=ink, anchor=south] at ($(top.south west)!0.5!(bot.north west)+(-0.92cm,0)$)
  {$\kappa=\Delta/C$ (add/remove)};
\draw[axisc, line width=0.45pt] ($(top.south west)+(-0.08cm,-0.03cm)$) -- ++(0.16cm,0.06cm);
\draw[axisc, line width=0.45pt] ($(bot.north west)+(-0.08cm,-0.03cm)$) -- ++(0.16cm,0.06cm);
\node[panel] at ($(top.north west)+(-1.0cm,0.12cm)$) {(a) Discovery: 64 candidate neighbors $\times$ 9 radii};
\begin{axis}[paper, name=val,
  at={($(top.north east)+(2.35cm,0)$)}, anchor=north west,
  width=3.75cm, height=4.16cm, scale only axis,
  xmin=-0.8, xmax=7.6, xtick={0,2,4,6}, xticklabel={$\pgfmathprintnumber{\tick}\%$},
  xmajorgrids,
  ymin=0.5, ymax=5.6, ytick={1,2,3,4,5},
  yticklabels={NF4 LoRA, BF16 LoRA, Threshold zero, Isolated records, Native batched},
  y tick label style={font=\scriptsize, text=ink},
  axis y line*=left, major tick length=0pt,
  xlabel={Excess over the bound, $\kappa-1$},
]
\draw[boundline] (axis cs:0,0.5) -- (axis cs:0,5.6);
\draw[native, line width=1.6pt] (axis cs:0,5) -- (axis cs:6.52579,5);
\foreach \y/\v in {4/-0.00015,3/-0.00024,2/-0.00015,1/-0.04702}{
  \edef\tmp{\noexpand\draw[isol, line width=1.6pt] (axis cs:0,\y) -- (axis cs:\v,\y);}\tmp}
\addplot[only marks, mark=*, mark size=2.2pt, native, mark options={draw=white, line width=0.6pt}] coordinates {(6.52579,5)};
\addplot[only marks, mark=*, mark size=2.2pt, isol, mark options={draw=white, line width=0.6pt}] coordinates {(-0.00015,4) (-0.00024,3) (-0.00015,2) (-0.04702,1)};
\node[note, anchor=north east, align=right] at (axis cs:7.5,4.55) {$+6.53\%$ at $C{=}0.25$\\ and $C{=}0.001$};
\node[note, anchor=west, align=left] at (axis cs:0.4,2.5) {all four controls\\ within $C$ at both\\ radii (to $-0.05\%$)};
\end{axis}
\node[panel] at ($(val.north west)+(-2.25cm,0.12cm)$) {(b) Native validation of the top pair};
\end{tikzpicture}
\caption{\textbf{Sensitivity on pretrained OPT-125M appears once clipping is active.} (a)~Exploratory add/remove ratio $\kappa=\Delta/C$ for all 64 candidate additions to one seven-record TinyStories batch, from re-clipping the saved native per-record gradients at nine radii (all 576 scores shown). Orange: the 10 candidates that exceed $C$ by more than $0.1\%$. At the original screening radius $C{=}1$ most gradients are not clipped, the added record contributes less than $C$, and no candidate exceeds the bound. (b)~Forty fresh native cases for the top candidate at $C{=}0.25$ and $C{=}0.001$: native batching exceeds the bound by $6.53\%$ at both radii, while record isolation, threshold zero, BF16 and NF4 stay within it. Identical-batch repeats give exactly zero distance, so the measurement floor is zero.}
\label{fig:landscape}
\end{figure*}
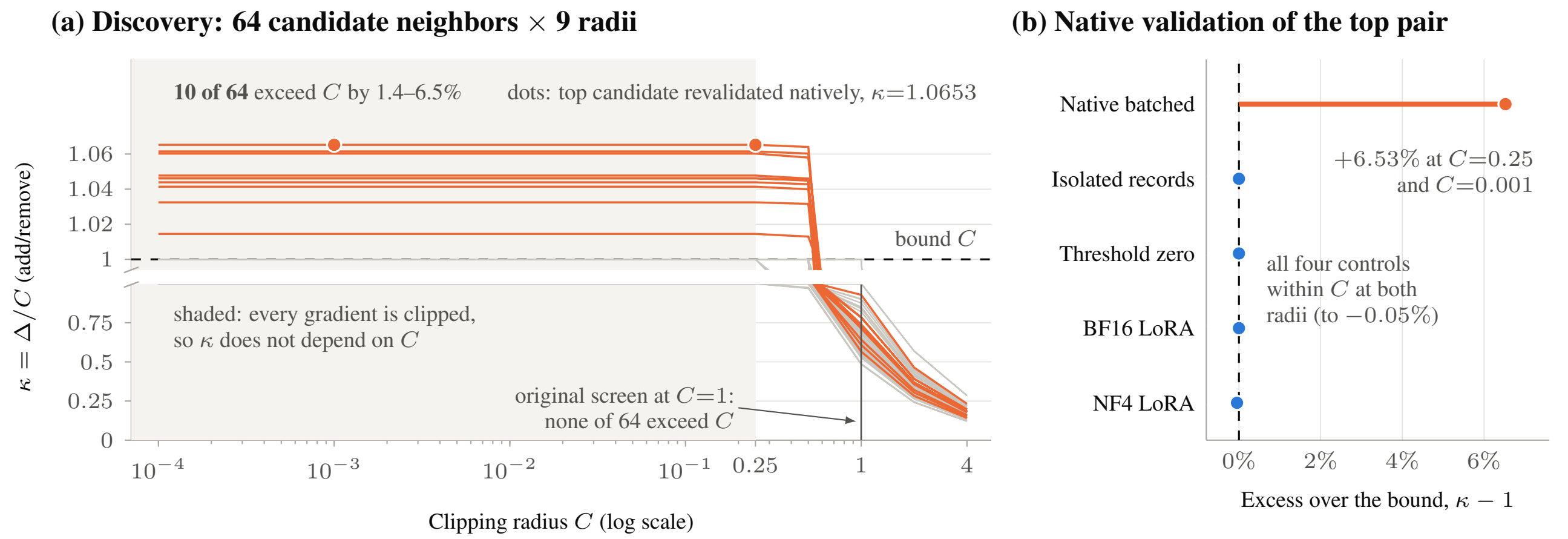

%% file: figures/fig_intervention.tex
\begin{figure*}[t]
\centering
\begin{tikzpicture}[font=\scriptsize]
\begin{scope}[x=1cm, y=1cm]
  \def\cw{0.5}\def\ch{0.46}
  \node[panel] at (-1.85,0.62) {(a) Which union changes? One of 72};
  \foreach \r/\lab in {0/{attn input}, 1/{attn output}, 2/{fc1}, 3/{fc2}}{
    \node[note, anchor=east, text=ink] at (-0.06,{-(\r+0.5)*\ch}) {\lab};}
  \pgfplotstableread[col sep=comma, row sep=crcr] {
row,module,layer,union_D,union_Dprime\\
0,attn input,0,0,0\\
0,attn input,1,0,0\\
0,attn input,2,6,6\\
0,attn input,3,6,6\\
0,attn input,4,8,8\\
0,attn input,5,5,5\\
0,attn input,6,6,6\\
0,attn input,7,7,7\\
0,attn input,8,4,4\\
0,attn input,9,3,3\\
0,attn input,10,3,3\\
0,attn input,11,3,3\\
1,attn output,0,0,0\\
1,attn output,1,0,0\\
1,attn output,2,0,0\\
1,attn output,3,0,0\\
1,attn output,4,0,0\\
1,attn output,5,0,0\\
1,attn output,6,0,0\\
1,attn output,7,0,0\\
1,attn output,8,0,0\\
1,attn output,9,0,0\\
1,attn output,10,0,0\\
1,attn output,11,0,0\\
2,fc1,0,0,0\\
2,fc1,1,4,4\\
2,fc1,2,2,2\\
2,fc1,3,1,1\\
2,fc1,4,1,1\\
2,fc1,5,1,1\\
2,fc1,6,1,2\\
2,fc1,7,1,1\\
2,fc1,8,2,2\\
2,fc1,9,2,2\\
2,fc1,10,2,2\\
2,fc1,11,6,6\\
3,fc2,0,8,8\\
3,fc2,1,0,0\\
3,fc2,2,0,0\\
3,fc2,3,0,0\\
3,fc2,4,0,0\\
3,fc2,5,0,0\\
3,fc2,6,0,0\\
3,fc2,7,0,0\\
3,fc2,8,0,0\\
3,fc2,9,0,0\\
3,fc2,10,0,0\\
3,fc2,11,0,0\\
}\uniontab
  \pgfplotstablegetrowsof{\uniontab}
  \pgfmathtruncatemacro{\lastrow}{\pgfplotsretval-1}
  \foreach \k in {0,...,\lastrow}{
    \pgfplotstablegetelem{\k}{row}\of\uniontab \let\rr\pgfplotsretval
    \pgfplotstablegetelem{\k}{layer}\of\uniontab \let\ll\pgfplotsretval
    \pgfplotstablegetelem{\k}{union_D}\of\uniontab \let\ua\pgfplotsretval
    \pgfplotstablegetelem{\k}{union_Dprime}\of\uniontab \let\ub\pgfplotsretval
    \pgfmathsetmacro{\xa}{\ll*\cw+0.022}\pgfmathsetmacro{\xb}{(\ll+1)*\cw-0.022}
    \pgfmathsetmacro{\ya}{-\rr*\ch-0.022}\pgfmathsetmacro{\yb}{-(\rr+1)*\ch+0.022}
    \ifnum\ua=\ub
      \ifnum\ua=0
        \draw[gridc, line width=0.4pt] (\xa,\ya) rectangle (\xb,\yb);
      \else
        \fill[wash] (\xa,\ya) rectangle (\xb,\yb);
        \node[font=\scriptsize, text=ink2] at ({(\ll+0.5)*\cw},{-(\rr+0.5)*\ch}) {\ua};
      \fi
    \else
      \fill[native] (\xa,\ya) rectangle (\xb,\yb);
      \node[font=\scriptsize, text=white, inner sep=0pt] at ({(\ll+0.5)*\cw},{-(\rr+0.5)*\ch}) {\ua{\tiny$\!\to\!$}\ub};
    \fi
  }
  \foreach \l in {0,...,11}{\node[note] at ({(\l+0.5)*\cw},{-4*\ch-0.16}) {\l};}
  \node[note, anchor=north] at ({6*\cw},{-4*\ch-0.28}) {decoder layer};
  \node[note, anchor=north, align=center] at ({6*\cw},{-4*\ch-0.62}) {cell: union size under $D$ (blank: none)};
\end{scope}
\begin{scope}[xshift=7.7cm, x=1cm, y=1cm]
  \def\cw{0.8}\def\ch{0.56}
  \node[panel] at (-1.08,0.62) {(b) Forward};
  \node[note, anchor=south] at (\cw,0.27) {values from};
  \node[note, anchor=south] at (0.5*\cw,-0.03) {$D$};
  \node[note, anchor=south] at (1.5*\cw,-0.03) {$D'$};
  \node[note, anchor=south east] at (-0.03,-0.03) {mask};
  \foreach \r/\lab in {0/{$O(D)$},1/{$O(D')$}}{
    \node[note, anchor=east] at (-0.03,{-(\r+0.5)*\ch}) {\lab};}
  \foreach \r/\c/\fillc/\txt/\tc in {0/0/wash/{$0$}/ink2, 0/1/wash/{$0$}/ink2,
                                     1/0/native/{$309.4$}/white, 1/1/native/{$309.4$}/white}{
    \fill[\fillc] ({\c*\cw+0.025},{-\r*\ch-0.025}) rectangle ({(\c+1)*\cw-0.025},{-(\r+1)*\ch+0.025});
    \node[font=\scriptsize\bfseries, text=\tc] at ({(\c+0.5)*\cw},{-(\r+0.5)*\ch}) {\txt};}
  \node[note, anchor=north, align=center] at (0.65*\cw,-2*\ch-0.08) {$\ell_2$ change in common-\\[-1pt]record logits vs.\ native $D$};
\end{scope}
\begin{axis}[paper, name=grad,
  at={(10.95cm,0.1cm)}, anchor=north west,
  width=3.5cm, height=1.95cm, scale only axis,
  ymode=log, ymin=5e-7, ymax=3,
  ytick={1e-6,1e-4,1e-2}, yminorticks=false,
  xmin=-0.32, xmax=1.32, xtick={0,1},
  xticklabels={native, forced $O(D')$},
  ymajorgrids,
]
\pgfplotstableread[col sep=comma, row sep=crcr] {
record,native_L2,forced_L2,reduction\\
16,0.07069414431364687,1.888606484860569e-06,37432\\
17,0.06164769344871308,2.0655935344303912e-06,29845\\
18,0.05802424931119786,1.465812450768778e-06,39585\\
19,0.04799178563016733,1.1485035775226063e-06,41786\\
20,0.06937864241195962,1.969482747117592e-06,35227\\
21,0.06792123373618142,2.678028222177986e-06,25362\\
22,0.05186731573341015,1.3587941728983054e-06,38172\\
}\maskgrad
\pgfplotstablegetrowsof{\maskgrad}
\pgfmathtruncatemacro{\lastrow}{\pgfplotsretval-1}
\foreach \k in {0,...,\lastrow}{
  \pgfplotstablegetelem{\k}{native_L2}\of\maskgrad \let\nv\pgfplotsretval
  \pgfplotstablegetelem{\k}{forced_L2}\of\maskgrad \let\fv\pgfplotsretval
  \edef\tmp{\noexpand\addplot[faint, line width=0.5pt, mark=none] coordinates {(0,\nv) (1,\fv)};}\tmp
}
\addplot[only marks, mark=*, mark size=1.8pt, native, mark options={draw=white, line width=0.4pt}] table[x expr=0, y=native_L2, col sep=comma, row sep=crcr] {
record,native_L2,forced_L2,reduction\\
16,0.07069414431364687,1.888606484860569e-06,37432\\
17,0.06164769344871308,2.0655935344303912e-06,29845\\
18,0.05802424931119786,1.465812450768778e-06,39585\\
19,0.04799178563016733,1.1485035775226063e-06,41786\\
20,0.06937864241195962,1.969482747117592e-06,35227\\
21,0.06792123373618142,2.678028222177986e-06,25362\\
22,0.05186731573341015,1.3587941728983054e-06,38172\\
};
\addplot[only marks, mark=*, mark size=1.8pt, isol, mark options={draw=white, line width=0.4pt}] table[x expr=1, y=forced_L2, col sep=comma, row sep=crcr] {
record,native_L2,forced_L2,reduction\\
16,0.07069414431364687,1.888606484860569e-06,37432\\
17,0.06164769344871308,2.0655935344303912e-06,29845\\
18,0.05802424931119786,1.465812450768778e-06,39585\\
19,0.04799178563016733,1.1485035775226063e-06,41786\\
20,0.06937864241195962,1.969482747117592e-06,35227\\
21,0.06792123373618142,2.678028222177986e-06,25362\\
22,0.05186731573341015,1.3587941728983054e-06,38172\\
};
\node[note, anchor=south west, align=left, inner sep=1pt] at (axis cs:-0.3,6.5e-7) {$2.5$--$4.2{\times}10^{4}{\times}$ smaller};
\node[note, anchor=north east, align=right, inner sep=1pt] at (axis cs:1.3,2.6) {7 common records};
\end{axis}
\node[panel] at (10.2,0.62) {(c) Backward};
\node[note, anchor=south west] at ($(grad.north west)+(-0.62cm,0.02cm)$) {$\|g_r(D')-g_r(D)\|_2$};
\node[note, anchor=north] at ($(grad.south)+(0,-0.42cm)$) {clipped sum: $\kappa=1.0653\rightarrow0.999998$};
\end{tikzpicture}
\caption{\textbf{One layer's outlier mask causes the excess.} Selected OPT-125M pair at $C{=}0.25$, with $D$ (7 records) and $D'$ (8 records). (a)~Of the 72 INT8 unions, 42 are nonempty and exactly one differs between $D$ and $D'$: decoder layer~6's first feed-forward projection gains one column. (b)~Forward factorial at that layer, input batch fixed to $D$: the common records' activations are bitwise identical under $D$ and $D'$, and swapping only the mask reproduces the entire logit change. (c)~Forcing that layer to use the frozen union $O(D')$ through native forward and backward shrinks every common record's gradient change by four orders of magnitude and brings the clipped sum below $C$. The residual is nonzero, so exact independence is not established, and the post hoc mask is a diagnostic rather than a private policy.}
\label{fig:intervention}
\end{figure*}
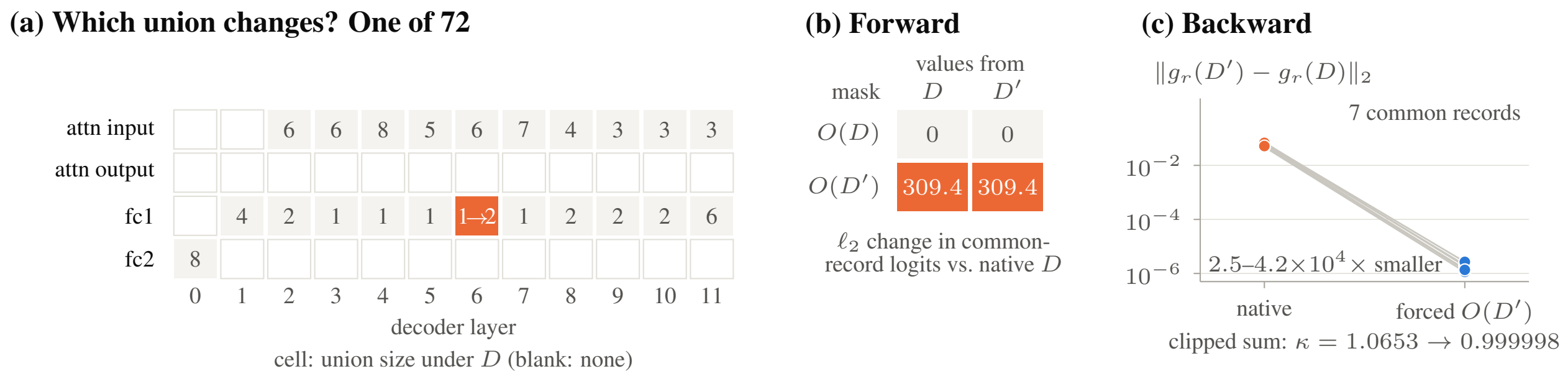

%% file: figures/fig_release.tex
\begin{figure}[t]
\centering
\begin{tikzpicture}
\begin{axis}[paper, name=rel,
  width=6.55cm, height=2.95cm, scale only axis,
  xmode=log, xmin=1e-7, xmax=1e-2,
  xtick={1e-7,1e-6,1e-5,1e-4,1e-3,1e-2},
  ymin=0, ymax=6.4, ytick={0,2,4,6}, ymajorgrids,
  xlabel={$\delta$}, ylabel={$\varepsilon$}, ylabel style={rotate=-90, yshift=-1pt},
  xlabel style={yshift=3pt},
  legend style={at={(0.985,0.02)}, anchor=south east, fill=white, inner sep=1.5pt, row sep=-2pt},
]
\addplot[name path=ideal, ink, line width=0.9pt] table[x=delta, y=eps_mu1, col sep=comma, row sep=crcr] {
delta,eps_mu1,eps_pair\\
1.000000e-07,5.349345,5.744049\\
1.211528e-07,5.312274,5.704621\\
1.467799e-07,5.274949,5.664923\\
1.778279e-07,5.237365,5.624952\\
2.154435e-07,5.199518,5.5847\\
2.610157e-07,5.161402,5.544163\\
3.162278e-07,5.123011,5.503335\\
3.831187e-07,5.084339,5.462209\\
4.641589e-07,5.045381,5.42078\\
5.623413e-07,5.00613,5.379039\\
6.812921e-07,4.96658,5.336982\\
8.254042e-07,4.926723,5.2946\\
1.000000e-06,4.886554,5.251886\\
1.211528e-06,4.846065,5.208832\\
1.467799e-06,4.805248,5.165431\\
1.778279e-06,4.764095,5.121675\\
2.154435e-06,4.722599,5.077554\\
2.610157e-06,4.680751,5.033061\\
3.162278e-06,4.638543,4.988185\\
3.831187e-06,4.595964,4.942917\\
4.641589e-06,4.553007,4.897247\\
5.623413e-06,4.50966,4.851164\\
6.812921e-06,4.465913,4.804658\\
8.254042e-06,4.421756,4.757718\\
1.000000e-05,4.377178,4.71033\\
1.211528e-05,4.332166,4.662484\\
1.467799e-05,4.286709,4.614165\\
1.778279e-05,4.240793,4.565361\\
2.154435e-05,4.194405,4.516056\\
2.610157e-05,4.14753,4.466237\\
3.162278e-05,4.100155,4.415886\\
3.831187e-05,4.052262,4.364988\\
4.641589e-05,4.003836,4.313525\\
5.623413e-05,3.954859,4.261479\\
6.812921e-05,3.905313,4.208831\\
8.254042e-05,3.855179,4.15556\\
1.000000e-04,3.804436,4.101644\\
1.211528e-04,3.753063,4.047062\\
1.467799e-04,3.701036,3.991788\\
1.778279e-04,3.648333,3.935798\\
2.154435e-04,3.594926,3.879064\\
2.610157e-04,3.540789,3.821557\\
3.162278e-04,3.485893,3.763247\\
3.831187e-04,3.430206,3.704101\\
4.641589e-04,3.373695,3.644084\\
5.623413e-04,3.316326,3.58316\\
6.812921e-04,3.25806,3.521287\\
8.254042e-04,3.198856,3.458422\\
1.000000e-03,3.138671,3.394521\\
1.211528e-03,3.077456,3.329533\\
1.467799e-03,3.015162,3.263404\\
1.778279e-03,2.951733,3.196076\\
2.154435e-03,2.887109,3.127486\\
2.610157e-03,2.821225,3.057566\\
3.162278e-03,2.754009,2.98624\\
3.831187e-03,2.685384,2.913427\\
4.641589e-03,2.615263,2.839036\\
5.623413e-03,2.543554,2.76297\\
6.812921e-03,2.470151,2.685117\\
8.254042e-03,2.394939,2.605358\\
1.000000e-02,2.317789,2.523556\\
};
\addlegendentry{ideal Gaussian, advertised sensitivity ($\mu{=}1$)}
\addplot[native, line width=0.9pt] table[x=delta, y=eps_pair, col sep=comma, row sep=crcr] {
delta,eps_mu1,eps_pair\\
1.000000e-07,5.349345,5.744049\\
1.211528e-07,5.312274,5.704621\\
1.467799e-07,5.274949,5.664923\\
1.778279e-07,5.237365,5.624952\\
2.154435e-07,5.199518,5.5847\\
2.610157e-07,5.161402,5.544163\\
3.162278e-07,5.123011,5.503335\\
3.831187e-07,5.084339,5.462209\\
4.641589e-07,5.045381,5.42078\\
5.623413e-07,5.00613,5.379039\\
6.812921e-07,4.96658,5.336982\\
8.254042e-07,4.926723,5.2946\\
1.000000e-06,4.886554,5.251886\\
1.211528e-06,4.846065,5.208832\\
1.467799e-06,4.805248,5.165431\\
1.778279e-06,4.764095,5.121675\\
2.154435e-06,4.722599,5.077554\\
2.610157e-06,4.680751,5.033061\\
3.162278e-06,4.638543,4.988185\\
3.831187e-06,4.595964,4.942917\\
4.641589e-06,4.553007,4.897247\\
5.623413e-06,4.50966,4.851164\\
6.812921e-06,4.465913,4.804658\\
8.254042e-06,4.421756,4.757718\\
1.000000e-05,4.377178,4.71033\\
1.211528e-05,4.332166,4.662484\\
1.467799e-05,4.286709,4.614165\\
1.778279e-05,4.240793,4.565361\\
2.154435e-05,4.194405,4.516056\\
2.610157e-05,4.14753,4.466237\\
3.162278e-05,4.100155,4.415886\\
3.831187e-05,4.052262,4.364988\\
4.641589e-05,4.003836,4.313525\\
5.623413e-05,3.954859,4.261479\\
6.812921e-05,3.905313,4.208831\\
8.254042e-05,3.855179,4.15556\\
1.000000e-04,3.804436,4.101644\\
1.211528e-04,3.753063,4.047062\\
1.467799e-04,3.701036,3.991788\\
1.778279e-04,3.648333,3.935798\\
2.154435e-04,3.594926,3.879064\\
2.610157e-04,3.540789,3.821557\\
3.162278e-04,3.485893,3.763247\\
3.831187e-04,3.430206,3.704101\\
4.641589e-04,3.373695,3.644084\\
5.623413e-04,3.316326,3.58316\\
6.812921e-04,3.25806,3.521287\\
8.254042e-04,3.198856,3.458422\\
1.000000e-03,3.138671,3.394521\\
1.211528e-03,3.077456,3.329533\\
1.467799e-03,3.015162,3.263404\\
1.778279e-03,2.951733,3.196076\\
2.154435e-03,2.887109,3.127486\\
2.610157e-03,2.821225,3.057566\\
3.162278e-03,2.754009,2.98624\\
3.831187e-03,2.685384,2.913427\\
4.641589e-03,2.615263,2.839036\\
5.623413e-03,2.543554,2.76297\\
6.812921e-03,2.470151,2.685117\\
8.254042e-03,2.394939,2.605358\\
1.000000e-02,2.317789,2.523556\\
};
\addlegendentry{ideal Gaussian, measured pair ($\mu{=}1.065$)}
\addplot[only marks, mark=triangle*, mark size=2.7pt, ink, mark options={draw=white, line width=0.4pt}]
  table[x=delta, y=eps_lb, col sep=comma, row sep=crcr] {
delta,eps_lb,eps_mu1,eps_pair,pos,neg,n\\
0.001,2.528549183638244,3.1386705485829403,3.3945211360275787,1043,23,500000\\
1e-05,3.250863213639772,4.377178095681225,4.71033040285613,1043,23,500000\\
1e-06,3.255528430945392,4.8865541174622145,5.251885614738914,1043,23,500000\\
};
\addlegendentry{MC lower bound from actual releases}
\path[name path=ceil] (axis cs:1e-7,6.4) -- (axis cs:1e-2,6.4);
\addplot[wash, forget plot] fill between[of=ideal and ceil];
\addplot[ink, line width=0.9pt, forget plot] table[x=delta, y=eps_mu1, col sep=comma, row sep=crcr] {
delta,eps_mu1,eps_pair\\
1.000000e-07,5.349345,5.744049\\
1.211528e-07,5.312274,5.704621\\
1.467799e-07,5.274949,5.664923\\
1.778279e-07,5.237365,5.624952\\
2.154435e-07,5.199518,5.5847\\
2.610157e-07,5.161402,5.544163\\
3.162278e-07,5.123011,5.503335\\
3.831187e-07,5.084339,5.462209\\
4.641589e-07,5.045381,5.42078\\
5.623413e-07,5.00613,5.379039\\
6.812921e-07,4.96658,5.336982\\
8.254042e-07,4.926723,5.2946\\
1.000000e-06,4.886554,5.251886\\
1.211528e-06,4.846065,5.208832\\
1.467799e-06,4.805248,5.165431\\
1.778279e-06,4.764095,5.121675\\
2.154435e-06,4.722599,5.077554\\
2.610157e-06,4.680751,5.033061\\
3.162278e-06,4.638543,4.988185\\
3.831187e-06,4.595964,4.942917\\
4.641589e-06,4.553007,4.897247\\
5.623413e-06,4.50966,4.851164\\
6.812921e-06,4.465913,4.804658\\
8.254042e-06,4.421756,4.757718\\
1.000000e-05,4.377178,4.71033\\
1.211528e-05,4.332166,4.662484\\
1.467799e-05,4.286709,4.614165\\
1.778279e-05,4.240793,4.565361\\
2.154435e-05,4.194405,4.516056\\
2.610157e-05,4.14753,4.466237\\
3.162278e-05,4.100155,4.415886\\
3.831187e-05,4.052262,4.364988\\
4.641589e-05,4.003836,4.313525\\
5.623413e-05,3.954859,4.261479\\
6.812921e-05,3.905313,4.208831\\
8.254042e-05,3.855179,4.15556\\
1.000000e-04,3.804436,4.101644\\
1.211528e-04,3.753063,4.047062\\
1.467799e-04,3.701036,3.991788\\
1.778279e-04,3.648333,3.935798\\
2.154435e-04,3.594926,3.879064\\
2.610157e-04,3.540789,3.821557\\
3.162278e-04,3.485893,3.763247\\
3.831187e-04,3.430206,3.704101\\
4.641589e-04,3.373695,3.644084\\
5.623413e-04,3.316326,3.58316\\
6.812921e-04,3.25806,3.521287\\
8.254042e-04,3.198856,3.458422\\
1.000000e-03,3.138671,3.394521\\
1.211528e-03,3.077456,3.329533\\
1.467799e-03,3.015162,3.263404\\
1.778279e-03,2.951733,3.196076\\
2.154435e-03,2.887109,3.127486\\
2.610157e-03,2.821225,3.057566\\
3.162278e-03,2.754009,2.98624\\
3.831187e-03,2.685384,2.913427\\
4.641589e-03,2.615263,2.839036\\
5.623413e-03,2.543554,2.76297\\
6.812921e-03,2.470151,2.685117\\
8.254042e-03,2.394939,2.605358\\
1.000000e-02,2.317789,2.523556\\
};
\addplot[native, line width=0.9pt, forget plot] table[x=delta, y=eps_pair, col sep=comma, row sep=crcr] {
delta,eps_mu1,eps_pair\\
1.000000e-07,5.349345,5.744049\\
1.211528e-07,5.312274,5.704621\\
1.467799e-07,5.274949,5.664923\\
1.778279e-07,5.237365,5.624952\\
2.154435e-07,5.199518,5.5847\\
2.610157e-07,5.161402,5.544163\\
3.162278e-07,5.123011,5.503335\\
3.831187e-07,5.084339,5.462209\\
4.641589e-07,5.045381,5.42078\\
5.623413e-07,5.00613,5.379039\\
6.812921e-07,4.96658,5.336982\\
8.254042e-07,4.926723,5.2946\\
1.000000e-06,4.886554,5.251886\\
1.211528e-06,4.846065,5.208832\\
1.467799e-06,4.805248,5.165431\\
1.778279e-06,4.764095,5.121675\\
2.154435e-06,4.722599,5.077554\\
2.610157e-06,4.680751,5.033061\\
3.162278e-06,4.638543,4.988185\\
3.831187e-06,4.595964,4.942917\\
4.641589e-06,4.553007,4.897247\\
5.623413e-06,4.50966,4.851164\\
6.812921e-06,4.465913,4.804658\\
8.254042e-06,4.421756,4.757718\\
1.000000e-05,4.377178,4.71033\\
1.211528e-05,4.332166,4.662484\\
1.467799e-05,4.286709,4.614165\\
1.778279e-05,4.240793,4.565361\\
2.154435e-05,4.194405,4.516056\\
2.610157e-05,4.14753,4.466237\\
3.162278e-05,4.100155,4.415886\\
3.831187e-05,4.052262,4.364988\\
4.641589e-05,4.003836,4.313525\\
5.623413e-05,3.954859,4.261479\\
6.812921e-05,3.905313,4.208831\\
8.254042e-05,3.855179,4.15556\\
1.000000e-04,3.804436,4.101644\\
1.211528e-04,3.753063,4.047062\\
1.467799e-04,3.701036,3.991788\\
1.778279e-04,3.648333,3.935798\\
2.154435e-04,3.594926,3.879064\\
2.610157e-04,3.540789,3.821557\\
3.162278e-04,3.485893,3.763247\\
3.831187e-04,3.430206,3.704101\\
4.641589e-04,3.373695,3.644084\\
5.623413e-04,3.316326,3.58316\\
6.812921e-04,3.25806,3.521287\\
8.254042e-04,3.198856,3.458422\\
1.000000e-03,3.138671,3.394521\\
1.211528e-03,3.077456,3.329533\\
1.467799e-03,3.015162,3.263404\\
1.778279e-03,2.951733,3.196076\\
2.154435e-03,2.887109,3.127486\\
2.610157e-03,2.821225,3.057566\\
3.162278e-03,2.754009,2.98624\\
3.831187e-03,2.685384,2.913427\\
4.641589e-03,2.615263,2.839036\\
5.623413e-03,2.543554,2.76297\\
6.812921e-03,2.470151,2.685117\\
8.254042e-03,2.394939,2.605358\\
1.000000e-02,2.317789,2.523556\\
};
\addplot[only marks, mark=triangle*, mark size=2.7pt, ink, mark options={draw=white, line width=0.4pt}, forget plot]
  table[x=delta, y=eps_lb, col sep=comma, row sep=crcr] {
delta,eps_lb,eps_mu1,eps_pair,pos,neg,n\\
0.001,2.528549183638244,3.1386705485829403,3.3945211360275787,1043,23,500000\\
1e-05,3.250863213639772,4.377178095681225,4.71033040285613,1043,23,500000\\
1e-06,3.255528430945392,4.8865541174622145,5.251885614738914,1043,23,500000\\
};
\node[note, anchor=north east, align=right] at (axis cs:8e-3,6.3) {an MC lower bound in this\\ region would refute the guarantee};
\node[note, anchor=east] at (axis cs:7.8e-7,3.26) {$3.26$};
\node[note, anchor=west] at (axis cs:1.3e-5,3.25) {$3.25$};
\node[note, anchor=south east, inner sep=1pt] at (axis cs:8e-4,2.6) {$2.53$};
\draw[ink2, line width=0.4pt, densely dotted] (axis cs:1e-5,3.42) -- (axis cs:1e-5,4.30);
\fill[ink] (axis cs:1e-5,4.37718) circle (0.9pt);
\node[note, anchor=north east] at (axis cs:0.85e-5,4.3) {$4.38$};
\end{axis}
\node[panel] at ($(rel.north west)+(-0.62cm,0.08cm)$) {(a) Measured bounds versus the guarantee};
\begin{axis}[paper, name=pow,
  at={($(rel.south west)+(0,-1.95cm)$)}, anchor=north west,
  width=6.55cm, height=2.35cm, scale only axis,
  xmode=log, xmin=1e5, xmax=1e11,
  xtick={1e5,1e6,1e7,1e8,1e9,1e10,1e11},
  ymin=2.6, ymax=5.15, ytick={3,3.5,4,4.5}, ymajorgrids,
  yticklabel style={/pgf/number format/fixed, /pgf/number format/precision=1},
  xlabel={Held-out draws per neighbor}, ylabel={expected $\varepsilon_{\rm lb}$},
  xlabel style={yshift=3pt},
]
\draw[boundline] (axis cs:1e5,4.37718) -- (axis cs:1e11,4.37718);
\addplot[ink, line width=0.9pt] table[x=n_heldout_per_side, y=eps_lb_mu1, col sep=comma, row sep=crcr] {
n_heldout_per_side,eps_lb_mu1,eps_lb_pair\\
1.000000e+05,2.7059,2.8954\\
1.467799e+05,2.8047,2.9994\\
2.154435e+05,2.9012,3.1012\\
3.162278e+05,2.9953,3.2005\\
4.641589e+05,3.0873,3.2976\\
6.812921e+05,3.1771,3.3925\\
1.000000e+06,3.2647,3.4853\\
1.467799e+06,3.3501,3.5758\\
2.154435e+06,3.4332,3.664\\
3.162278e+06,3.5139,3.7497\\
4.641589e+06,3.592,3.8331\\
6.812921e+06,3.6672,3.9136\\
1.000000e+07,3.7394,3.9912\\
1.467799e+07,3.808,4.0654\\
2.154435e+07,3.8726,4.1358\\
3.162278e+07,3.9333,4.2021\\
4.641589e+07,3.9893,4.264\\
6.812921e+07,4.0406,4.3211\\
1.000000e+08,4.087,4.3731\\
1.467799e+08,4.1284,4.4201\\
2.154435e+08,4.1651,4.462\\
3.162278e+08,4.1973,4.499\\
4.641589e+08,4.2251,4.5313\\
6.812921e+08,4.2493,4.5592\\
1.000000e+09,4.2699,4.5832\\
1.467799e+09,4.2873,4.6039\\
2.154435e+09,4.3023,4.6211\\
3.162278e+09,4.3148,4.636\\
4.641589e+09,4.3251,4.6485\\
6.812921e+09,4.334,4.6587\\
1.000000e+10,4.3414,4.6675\\
1.467799e+10,4.3475,4.6749\\
2.154435e+10,4.3526,4.681\\
3.162278e+10,4.3567,4.686\\
4.641589e+10,4.3602,4.6901\\
6.812921e+10,4.363,4.6936\\
1.000000e+11,4.3655,4.6964\\
};
\addplot[native, line width=0.9pt] table[x=n_heldout_per_side, y=eps_lb_pair, col sep=comma, row sep=crcr] {
n_heldout_per_side,eps_lb_mu1,eps_lb_pair\\
1.000000e+05,2.7059,2.8954\\
1.467799e+05,2.8047,2.9994\\
2.154435e+05,2.9012,3.1012\\
3.162278e+05,2.9953,3.2005\\
4.641589e+05,3.0873,3.2976\\
6.812921e+05,3.1771,3.3925\\
1.000000e+06,3.2647,3.4853\\
1.467799e+06,3.3501,3.5758\\
2.154435e+06,3.4332,3.664\\
3.162278e+06,3.5139,3.7497\\
4.641589e+06,3.592,3.8331\\
6.812921e+06,3.6672,3.9136\\
1.000000e+07,3.7394,3.9912\\
1.467799e+07,3.808,4.0654\\
2.154435e+07,3.8726,4.1358\\
3.162278e+07,3.9333,4.2021\\
4.641589e+07,3.9893,4.264\\
6.812921e+07,4.0406,4.3211\\
1.000000e+08,4.087,4.3731\\
1.467799e+08,4.1284,4.4201\\
2.154435e+08,4.1651,4.462\\
3.162278e+08,4.1973,4.499\\
4.641589e+08,4.2251,4.5313\\
6.812921e+08,4.2493,4.5592\\
1.000000e+09,4.2699,4.5832\\
1.467799e+09,4.2873,4.6039\\
2.154435e+09,4.3023,4.6211\\
3.162278e+09,4.3148,4.636\\
4.641589e+09,4.3251,4.6485\\
6.812921e+09,4.334,4.6587\\
1.000000e+10,4.3414,4.6675\\
1.467799e+10,4.3475,4.6749\\
2.154435e+10,4.3526,4.681\\
3.162278e+10,4.3567,4.686\\
4.641589e+10,4.3602,4.6901\\
6.812921e+10,4.363,4.6936\\
1.000000e+11,4.3655,4.6964\\
};
\addplot[only marks, mark=triangle*, mark size=2.7pt, ink, mark options={draw=white, line width=0.4pt}] coordinates {(5e5,3.250863)};
\draw[ink2, line width=0.4pt, densely dotted] (axis cs:1.032e8,2.6) -- (axis cs:1.032e8,4.37718);
\node[note, anchor=north west] at (axis cs:1.15e5,4.33) {advertised $\varepsilon{=}4.38$};
\node[note, anchor=north west, align=left, inner sep=1pt] (aud) at (axis cs:2.2e6,3.02) {our audit: $3.25$ ($5{\times}10^5$ per neighbor)};
\draw[ink2, line width=0.35pt] (aud.west) -- (axis cs:6.2e5,3.2);
\node[note, anchor=south west, align=left, inner sep=1pt] (need) at (axis cs:1.15e5,4.52) {$\approx10^{8}$ draws per neighbor\\ needed to resolve $\mu{=}1.065$};
\draw[-{Latex[length=1.2mm]}, ink2, line width=0.35pt] (need.east) -- (axis cs:0.85e8,4.43);
\node[note, anchor=south east] at (axis cs:9e10,4.72) {$\mu{=}1.065$};
\node[note, anchor=north east] at (axis cs:9e10,4.27) {$\mu{=}1$};
\end{axis}
\node[panel] at ($(pow.north west)+(-0.62cm,0.08cm)$) {(b) How large an audit would need to be};
\end{tikzpicture}
\caption{\textbf{No release-level violation is detected, and none could be at this audit size.} (a)~Two million actual FP32 adapter releases of the selected pair ($10^6$ per neighbor). Triangles: held-out Monte Carlo lower bounds on $\varepsilon$ (family-wise 95\% Clopper--Pearson). Lines: exact real-valued Gaussian privacy curves (Eq.~\ref{eq:gaussian}) at the advertised sensitivity (black) and at the measured pair (orange). (b)~Expected bound at $\delta{=}10^{-5}$ when the same procedure audits an ideal Gaussian release, with an oracle threshold (analytic). Our observed bound lies between the two curves; a bound above the advertised $\varepsilon$ needs roughly 200 times more draws.}
\label{fig:release}
\end{figure}
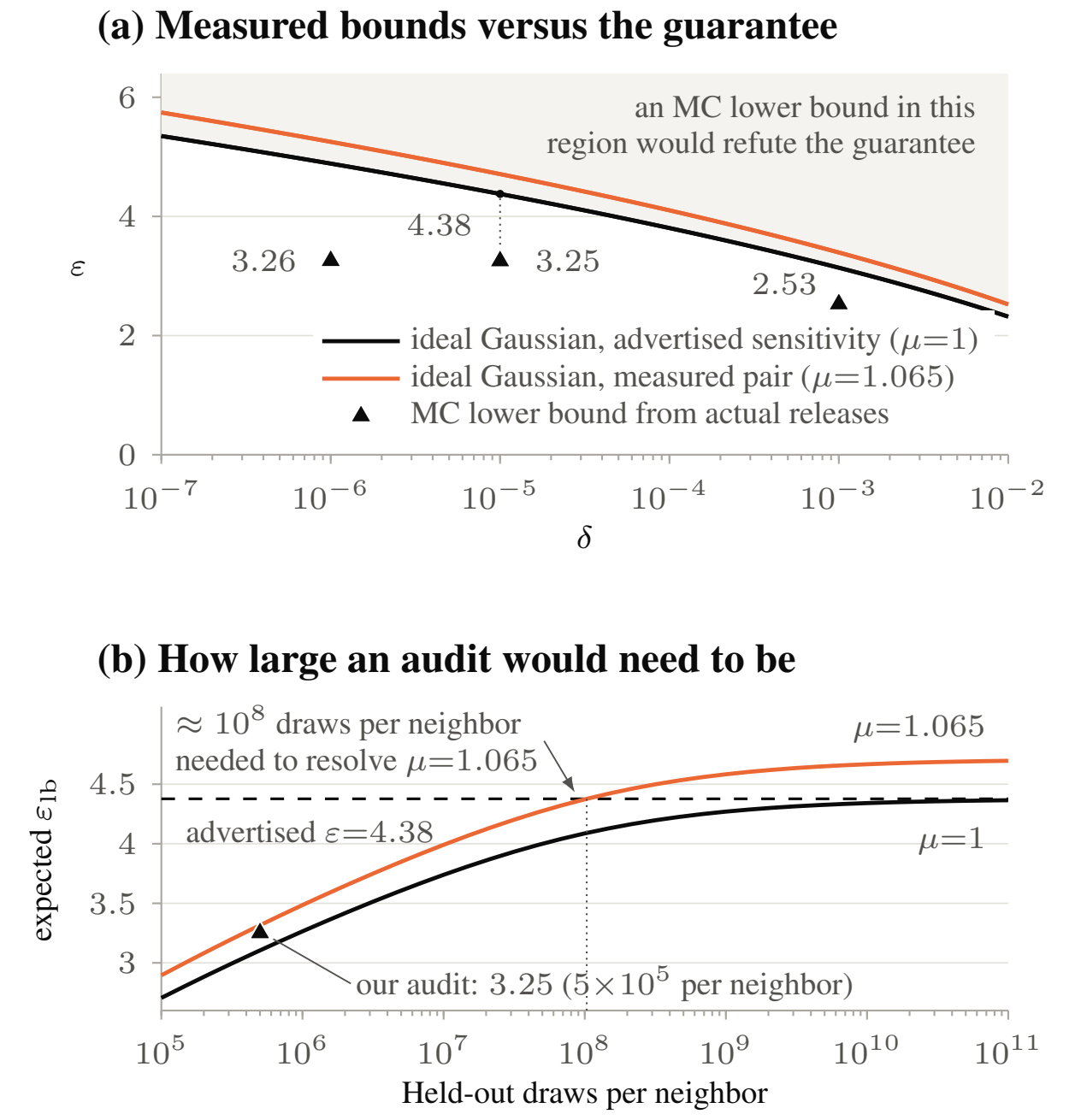

%% file: appendix.tex
\section{Constructed Witness Details}
\label{app:construction}
\paragraph{Operator arithmetic.} The frozen layer is a genuine bitsandbytes \texttt{Linear8bitLt} with frozen \texttt{Int8Params} and no internal bias; an external FP32 scalar bias is the only trainable parameter. The weight $[1,0]$ and all activation rows are padded with two exact zeros to inner dimension $K{=}4$. Each row's scale is its absolute maximum over entries below $\tau{=}6$, which is 5 for every row in both batches. Without an outlier column, the unchanged entry $1.234$ is coded as $\mathrm{round}(1.234\cdot127/5)=31$ and dequantizes to $31\cdot5/127\approx1.2205$, stored in FP16 as $1.220703125$. Once the replaced row's $6.1$ puts column~0 into the union, the same entry is zeroed in the INT8 codes and recomputed in FP16, giving $1.234375$. The recorded native outputs equal these values exactly.

\paragraph{Labels and gradients.} A pilot obtains the two native unchanged-record outputs; their FP32 midpoint, $1.2275390625$, is then fixed as the common label of the seven unchanged records in both neighboring batches and in every control. With squared-error loss the unchanged records' bias derivatives are $-0.0068359375$ and $+0.0068359375$, both larger than $C$ in magnitude, so each clipped contribution moves from $-C$ to $+C$. The replaced record's label is $-100$, so its clipped contribution is $+C$ in both batches. The labels are common to the compared datasets but deliberately chosen; they do not represent a natural task.

\paragraph{Exact values.} Writing $C_f=0.004999999888241291$ for the FP32 clipping constant, the native replacement distance is $\Delta=14C_f=0.06999999843537807$, so $\Delta/(2C_f)$ is exactly 7. FP64 accumulation of the FP32 contributions is exact for this fixture, and the native FP32 sums agree, so the effect is not produced by a decimal clipping constant or a reduction discrepancy. The LoRA--Opacus composition uses summed losses with public divisor 1, $q{=}1$, $C{=}0.005$ and noise multiplier 1; its add/remove and replacement distances are $0.07499169931$ and $0.06999170035$. CPU or XPU dispatch can differ from CUDA, so only native CUDA executions count as evidence.

\section{Experimental Records and Provenance}
\label{app:provenance}
\paragraph{Records.} Every reported number traces to a machine-readable result record that stores the experiment source digest, package-source digests, the loaded CUDA library digest, the inputs and all case traces. The plotted values in Figures~\ref{fig:coupling}--\ref{fig:release} are derived from these records by a standard-library script, and the resolution estimate in \S\ref{sec:res-release} uses exact binomial tails; both scripts accompany the artifact.

\paragraph{Retained failures.} We report failed attempts alongside the accepted ones. The first OPT-125M reachability run rejected an incorrect 128-token shape assumption before loading the model; the accepted run uses the full 256-token windows. In the $C{=}1$ screen, four BF16 control attempts failed their exact-adapter initialization check before measurement; a separately versioned supplement loaded the canonical FP32 adapter bytes and completed exactly those four controls, so the screen comprises 86 measured conditions and four retained rejections rather than 90 independent observations. The gradient-level mask intervention exists in an original and a corrected version, and both are retained together with the CPU validation of the accepted one.

\paragraph{Independent reconciliation.} All raw arrays of the 40 selected-radius native cases were reconciled on CPU in FP64, reproducing every reported distance. For the release audit, the 1,024-draw-per-side qualification runs, which reproduce both parent releases bitwise, are excluded from the reported counts. A CPU pass verified all eight shard identities, seeds, scalar hashes, shapes and the complete two-million-draw coverage.